\documentclass[journal,10pt,twocolumn,twoside]{IEEEtran}
\usepackage{ifpdf}
\usepackage{cite}
\usepackage[table]{xcolor}
\usepackage{amsmath,amssymb,amsfonts,bm}
\usepackage{algorithmic}
\usepackage{graphicx, adjustbox}
\usepackage{array}
\usepackage{enumitem}
\usepackage{breqn}
\usepackage{bm}
\usepackage{ dsfont }
\usepackage{ upgreek }
\usepackage{textcomp}
\usepackage{multirow}
\usepackage{algorithm}
\usepackage{algorithmic}
\usepackage{caption}
\usepackage{pgfplots}
\usepackage{tkz-euclide}
\usepackage{subcaption}
\usepackage{tikz}
\usepackage{hyperref}
\usepackage{balance}
\usepackage{booktabs}
\usepackage{ tipa }

\usepackage{ upgreek }
\usepackage{soul}
\usepackage{breqn}
\usepackage{makecell}
\usepackage{tabularray}
\usepackage[english]{babel}

\usepackage{booktabs} 
\usepackage{amssymb} 
\usepackage{pifont}  
\usepackage{graphicx} 
\usepackage{adjustbox}
\usepackage{tabularx}

\addto\captionsenglish{}
\allowdisplaybreaks

\usepackage{amsthm}

\usepackage{fancyhdr}

\usepackage[most]{tcolorbox}
\usepackage{graphicx}
\usepackage{array}
\usepackage{makecell}
\usepackage{multirow}

\usepackage{graphicx}
\usepackage{adjustbox}
\usepackage{multirow}
\usepackage{makecell}
\usepackage{ upgreek }
\usepackage{soul}
\makeatletter
\newcommand{\vast}{\bBigg@{4}}

\newcommand{\Vast}{\bBigg@{5}}
\makeatother
\pgfplotsset{compat=1.18}
\usepackage{array}
\usepackage{makecell} 
\usepackage{amsmath}

\usepackage{booktabs}
\usepackage{amssymb}
\usepackage{pifont}
\usepackage{graphicx}
\usepackage{adjustbox}
\usepackage{tabularx} 

\newcommand{\cmark}{\ding{51}}
\newcommand{\xmark}{\ding{55}}
\newcommand{\partialmark}{\resizebox{!}{0.12in}{$\circ$}} 

\begin{document}
\title{Quantum Federated Learning: \\ A Comprehensive Survey}

\author{Dinh C. Nguyen, \textit{Member}, \textit{IEEE}, Md Raihan Uddin, Shaba Shaon, Ratun Rahman, \\ Octavia Dobre, \textit{Fellow}, \textit{IEEE}, and Dusit Niyato, \textit{Fellow}, \textit{IEEE}

\thanks{Dinh C. Nguyen, Md Raihand Uddin, Shaba Shaon, and Ratun Rahman are with the Department of Electrical and Computer Engineering, The University of Alabama in Huntsville, USA (e-mails: dinh.nguyen@uah.edu, mu0016@uah.edu, ss0670@uah.edu, rr0110@uah.edu).}
\thanks{Octavia Dobre is with the Faculty of Engineering and Applied Science, Memorial University, Canada (e-mail: odobre@mun.ca).}
\thanks{Dusit Niyato is with the College of Computer Science and Engineering, Nanyang Technological University, Singapore (e-mail: dniyato@ntu.edu.sg).}

}




\markboth{IEEE Communications Surveys \& Tutorials}%
	{}
\maketitle

\begin{abstract}
Quantum federated learning (QFL) is a combination of distributed quantum computing and federated machine learning, integrating the strengths of both to enable privacy-preserving decentralized learning with quantum-enhanced capabilities. It appears as a promising approach for addressing challenges in efficient and secure model training across distributed quantum systems. This paper presents a comprehensive survey on QFL, exploring its key concepts, fundamentals, applications, and emerging challenges in this rapidly developing field. Specifically, we begin with an introduction to the recent advancements of QFL, followed by discussion on its market opportunity and background knowledge. We then discuss the motivation behind the integration of quantum computing and federated learning, highlighting its working principle. Moreover, we review the fundamentals of QFL and its taxonomy. Particularly, we explore federation architecture, networking topology, communication schemes, optimization techniques, and security mechanisms within QFL frameworks. Furthermore, we investigate applications of QFL across several domains which include vehicular networks, healthcare networks, satellite networks, metaverse, and network security. Additionally, we analyze frameworks and platforms related to QFL, delving into its prototype implementations, and provide a detailed case study. Key insights and lessons learned from this review of QFL are also highlighted. We complete the survey by identifying current challenges and outlining potential avenues for future research in this rapidly advancing field. 
\end{abstract}



%
\IEEEpeerreviewmaketitle





\section {Introduction} \label{introduction}
Quantum computing holds the potential to revolutionize computation by offering exponential speedups for certain classes of problems that are intractable for classical computers. This promise has attracted considerable attention from scientific, industrial, and governmental communities. However, the current state of quantum hardware, known as the Noisy Intermediate-Scale Quantum (NISQ) era, is characterized by devices that have limited qubit counts, short coherence times, and high error rates. As a result, many complex quantum applications, such as quantum chemistry simulations—remain out of reach. These applications often require hundreds or even thousands of qubits to represent complex molecular systems accurately, as demonstrated in recent studies \cite{blunt2024compilation, jones2012faster}, far exceeding the capabilities of existing NISQ systems.

To address these limitations, Distributed Quantum Computing (DQC) has emerged as a promising architectural approach \cite{caleffi2024distributed, cuomo2023optimized, cuomo2020towards}. DQC envisions a network of interconnected small- to moderate-scale quantum processors that collaboratively execute computational tasks. Rather than relying on a single large quantum computer, DQC leverages modular quantum devices, potentially located across different physical locations, and links them through quantum communication channels. This paradigm enables the aggregation of quantum resources, effectively scaling the number of usable qubits and allowing for more sophisticated quantum algorithms to be implemented under the constraints of NISQ hardware.

Within this DQC framework, quantum federated learning (QFL) has emerged as a key enabling paradigm. QFL integrates principles of quantum computing with federated machine learning, allowing multiple quantum devices to jointly train QML models without sharing raw quantum data \cite{qiao2024transitioning}. This approach not only enhances privacy and data locality but also aligns naturally with the distributed structure of DQC. By coordinating the training process across various NISQ devices, QFL makes it feasible to tackle large-scale data processing tasks and QML workloads that would otherwise exceed the capacity of individual quantum processors \cite{tomar2025comprehensive}. Moreover, QFL introduces a scalable mechanism to effectively distribute the computational load and improve resource utilization in quantum networks. As the demand for quantum-enhanced machine learning grows across domains such as scientific computing, drug discovery, cybersecurity, and financial modeling, the importance of scalable and collaborative quantum intelligence becomes increasingly evident. QFL represent a forward-looking solution that bridges the current hardware limitations of the NISQ era and opens pathways for practical, near-term quantum applications \cite{gurung2023decentralized}.

\begin{table}[htbp]
    \centering
    \caption{List of key Acronyms.}
    \label{tab:qfl_acronyms}
    \small
    \begin{adjustbox}{max width=\textwidth}
    \begin{tabular}{|l|l|}
    \hline
    \textbf{Acronym} & \textbf{Definition} \\ \hline
    AI   & Artificial Intelligence             \\ \hline
    BQC  & Blind Quantum Computing             \\ \hline
    CQFL & Centralized Quantum Federated Learning \\ \hline
    DQFL & Decentralized Quantum Federated Learning \\ \hline
    FL   & Federated Learning                  \\ \hline
    FSO  & Free Space Optical                  \\ \hline
    HQFL & Hybrid Quantum Federated Learning   \\ \hline
    HQKD & Hierarchical Quantum Key Distribution \\ \hline
    IoT  & Internet of Things          \\ \hline
    IoQT & Internet of Quantum Things          \\ \hline
    ML   & Machine Learning                   \\ \hline
    NN   & Neural Network                   \\ \hline
    NISQ & Noisy Intermediate-Scale Quantum    \\ \hline
    P2P  & peer-to--peer                    \\ \hline
    PQC  & Parameterized Quantum Circuit       \\ \hline
    QC   & Quantum Computing \\ \hline
    QCNN & Quantum Convolutional Neural Network \\ \hline
    QCC  & Quantum-Classical Computing         \\ \hline
    QDP  & Quantum Differential Privacy        \\ \hline
    QEC  & Quantum Error Correction            \\ \hline
    QECC & Quantum Error Correction Codes      \\ \hline
    QED  & Quantum Error Detection             \\ \hline
    QFL  & Quantum Federated Learning          \\ \hline
    QML  & Quantum Machine Learning          \\ \hline
    QHE  & Quantum Homomorphic Encryption      \\ \hline
    QKD  & Quantum Key Distribution            \\ \hline
    QML  & Quantum Machine Learning            \\ \hline
    QNN  & Quantum Neural Network              \\ \hline
    QSN  & Quantum Satellite Networks          \\ \hline
    QVQE & Quantum Variational Quantum Eigensolver \\ \hline
    QVN  & Quantum Vehicular Networks          \\ \hline
    UAV  & Unmanned Aerial Vehicle          \\ \hline
    VQA  & Variational Quantum Algorithm       \\ \hline
    VQC  & Variational Quantum Circuit         \\ \hline
    VQE  & Variational Quantum Eigensolver      \\ \hline
    \end{tabular}
    \end{adjustbox}
\end{table}

\subsection{State-of-the-Art and Our Contributions}

A number of studies have explored the landscape of QFL, summarized in Table~\ref{tab:qfl_comparison} highlighting key developments, challenges, and future directions in this emerging field. 
\color{black}For instance, \cite{ren2025toward, ren2023towards, caleffi2024distributed} provide a valuable survey of the challenges and opportunities in the emerging field of QFL from a computer and communications engineering perspective. Their work effectively outlines the key findings and future directions for the field. While acknowledging this important contribution, our survey provides a distinct and complementary direction by offering a more granular and structured approach. Specifically, our work differentiates itself by presenting a comprehensive, multi-layered \textit{taxonomy} of QFL systems, covering architectures, networking topologies, communication schemes, and optimization strategies in greater depth. Furthermore, we provide a dedicated analysis of \textit{market and industry developments} and include a practical \textit{case study} to demonstrate the implementation of QFL with current frameworks. We design this approach not only to provide a high-level overview but also to equip researchers and practitioners with a detailed, foundational understanding and a guide to tangible implementation, thereby filling a critical gap in the existing literature. 
\color{black} The researchers in \cite{cacciapuoti2019quantum, yang2023survey} offered an introduction to quantum computing and its effects on the field of communications, reviewing key advancements and essential elements of quantum internet. They also investigated the challenges of designing the quantum internet and discussed key research directions in quantum network design, illustrating the shift required from classical network paradigms to accommodate quantum-specific phenomena, i.e., entanglement and quantum teleportation. The study in \cite{javeed2024quantum} investigated the integration of quantum computing, FL, and 6G networks and their role in enhancing IoT privacy and security. It also sheds light on the recent advancements in this field, while offering a design of a conceptual framework and outlining future implications for the field. \textit{However, authors have limited discussion on the applications of QFL in IoT systems.} An overview of the journey from FL to QFL was discussed in \cite{qiao2024transitioning}, providing basic concepts related to this transition. Authors highlighted integration strategies, current challenges, and future directions in leveraging quantum computing to enhance ML computational efficiency and security, \textit{overlooking several other aspects of QFL in their discussion}. Similarly, the discussion in \cite{chehimi2023foundations} was confined to QFL's key components, challenges, and solutions for deployment over classical and quantum networks. \textit{This work provided only a brief overview without delving into the holistic technical fundamentals as well as applications of QFL.} Another work \cite{wang2023quantum} proposed a quantum-empowered FL framework for space-air-ground integrated networks, employing variational quantum algorithms and quantum relays to enhance model training and secure long-distance model transmission. \textit{Nevertheless, this paper limited its discussion to the application of QFL within aerial networks only.} The study in \cite{nguyen2021federated} presented a comprehensive survey of FL applications within IoT networks, exploring how FL integrates and enhances various services. The studies in \cite{li2021survey, abdulrahman2020survey} conducted a review of FL systems, analyzing key system components and presenting design factors and studies. The authors compared different ML deployment architectures, offering a new classification of FL research, and discussing comprehensive taxonomies, challenges, and future directions in the field. \textit{Nonetheless, they did not explore the integration of quantum mechanics into the distributed ML environment.} 
\color{black}
In addition to these works, other recent surveys have also explored the QFL landscape. Ballester et al. \cite{ballester2025quantum} provide a comprehensive literature review covering the foundations, challenges, and future directions of QFL, while Saha et al. \cite{saha2024quantum} offer a concise overview of the progress and opportunities presented at a recent conference. These papers make valuable contributions to consolidating knowledge in this rapidly evolving field. Our survey builds upon and complements these efforts by providing several distinct contributions: 1) a more granular, multi-layered taxonomy that systematically classifies QFL architectures, communication schemes, and security mechanisms; 2) a unique focus on industry developments and commercialization prospects, bridging the gap between academic research and market potential; and 3) a practical case study that offers a tangible demonstration of QFL implementation using current tools. By integrating these theoretical, practical, and industry-focused perspectives, our work provides a holistic, end-to-end resource for the research community.
\color{black}

\textcolor{black}{While the recent field of QFL has received significant attention, existing surveys have primarily focused on its quantum-centric aspects, including novel circuit designs, quantum-enhanced security protocols, and theoretical performance gains. However, a critical gap remains in contextualizing QFL within the broader landscape of practical FL challenges. Foundational issues that govern the performance and feasibility of classical FL systems are often overlooked, yet they will undoubtedly persist and evolve in the quantum paradigm. Two such pivotal challenges are the management of FL processing on resource-constrained edge devices and the strategic selection of clients for training rounds \cite{ballester2025quantum}. Efficiently orchestrating model training at the edge requires sophisticated resource management to handle the trade-offs between communication, computation, and energy, as thoroughly surveyed in works like \cite{trindade2021management}. Similarly, client selection has evolved into a complex task where modern strategies must balance model convergence speed with system fairness and data heterogeneity \cite{tan2022reputation}. Neglecting these operational realities can render even the most advanced QFL algorithms impractical for real-world deployment.}

\textcolor{black}{To bridge this gap and provide a more holistic perspective, our survey makes a distinct contribution by integrating the principles of QFL with a grounded discussion of these persistent operational challenges. Unlike prior works that treat QFL in a theoretical aspect, we explicitly address how the unique constraints of edge computing will influence the design and deployment of future QFL networks.}

Although FL has been extensively explored in existing literature, and works have addressed various facets of QFL, \textit{no study to date offers a comprehensive review of QFL that includes its fundamental principles, taxonomy, applications, case studies, as well as market opportunities}. Motivated by these limitations, we conduct a comprehensive review of QFL. Particularly, we provide a state-of-the-art survey of QFL, starting from the fundamental knowledge behind this integration and extending our investigation to its applications in diverse fields. We explore the taxonomy of QFL along the way, from networking, communication, optimization, and security perspectives. We also present a case study of QFL, along with challenges and future research directions in this emerging yet promising field. Key insights and lessons learned are also presented. The key contributions of this work are highlighted as follows:
\begin{itemize}
    \item We present a state-of-the-art survey on QFL, starting with the basics of quantum computing and progressively expanding our exploration to include recent advancements in QFL, as well as a discussion on the vision and prospects that have driven its development.
    \item We discuss the industry and commercialization opportunities of QFL and their potential impact on future technological landscapes.
    \item We illustrate how QFL solves the constraints of traditional FL in terms of scalability, accuracy, and security, thereby offering a thorough study of the requirement of integrating quantum computing with FL. This comparison helps researchers move more successfully from FL to QFL-based solutions.
    \item We meticulously explore the taxonomy of QFL, detailing diverse federation architectures, networking topologies, communication schemes, optimization strategies, and security mechanisms to provide a foundational understanding of the advancements in QFL.
    \item We perform a holistic investigation and analysis of the potential application of QFL across various domains, such as vehicular networks, healthcare systems, satellite networks, the metaverse, and network security.
    \item We present a case study that demonstrates the practical application and effectiveness of QFL using current frameworks, platforms, and prototype applications.
    \item The key lessons learned are highlighted at the end of each section. Lastly, we identify a couple of important research challenges and then discuss possible directions for future research in the field of QFL.
\end{itemize}

\begin{table*}[h!]
    \color{black}
    \centering
    \caption{A Structured Comparison of Our Survey with Existing Literature Based on Key Coverage Areas.}
    \label{tab:qfl_comparison}
    \small
    \begin{adjustbox}{max width=\textwidth}
    \begin{tabularx}{\textwidth}{|l|X|X|X|c|c|c|c|c|}
        \hline
        \textbf{Ref.} & \textbf{Primary Focus} & \textbf{Key Contributions} & \textbf{Limitations} & \textbf{Principles} & \textbf{Taxonomy} & \textbf{Apps} & \textbf{Case Studies} & \textbf{Market} \\ \hline

        \cite{ren2025toward} & QFL Concepts & Challenges, opportunities, engineering perspective. & Limited scope on basics, networking, and applications. & \partialmark & \cmark & \partialmark & \xmark & \xmark \\ \hline
        
        \cite{ballester2025quantum} & QFL Literature Review & Foundations, challenges, future directions. & Lacks deep taxonomy and implementation details. & \cmark & \partialmark & \cmark & \xmark & \xmark \\ \hline
        
        \cite{saha2024quantum} & QFL Progress & Conference overview of progress, challenges. & High-level; not an in-depth survey. & \partialmark & \xmark & \partialmark & \xmark & \xmark \\ \hline
        
        \cite{caleffi2024distributed} & Quantum Internet & Quantum network design, entanglement. & Lacks comprehensive QFL principles and applications. & \partialmark & \xmark & \partialmark & \xmark & \xmark \\ \hline
        
        \cite{qiao2024transitioning} & FL to QFL Transition & Integration strategies, challenges, efficiency. & Overlooks broad discussion of applications. & \cmark & \partialmark & \xmark & \xmark & \xmark \\ \hline

        \cite{chehimi2023foundations} & QFL Components & Key components, deployment challenges. & Brief overview; lacks deep technical fundamentals. & \partialmark & \cmark & \partialmark & \xmark & \xmark \\ \hline

        \cite{javeed2024quantum} & QFL in 6G/IoT & Integration with 6G/IoT, privacy, security. & Limited discussion on broader QFL applications. & \partialmark & \partialmark & \cmark & \xmark & \xmark \\ \hline

        \cite{wang2023quantum} & QFL in Aerial Nets & Space-air-ground networks, VQAs. & Scope limited to the single application of aerial networks. & \partialmark & \xmark & \cmark & \xmark & \xmark \\ \hline

        \cite{nguyen2021federated} & Classical FL in IoT & FL applications in IoT, challenges. & Does not explore quantum computing integration. & \xmark & \cmark & \cmark & \partialmark & \xmark \\ \hline

        \cite{li2021survey} & Classical FL Systems & System components, design factors, case studies. & Does not delve into quantum mechanics integration. & \xmark & \cmark & \partialmark & \cmark & \xmark \\ \hline

        \cite{abdulrahman2020survey} & Classical FL Archs. & ML deployment architectures, FL classification. & Overlooks quantum mechanics integration. & \xmark & \cmark & \xmark & \partialmark & \xmark \\ \hline
        
        \textit{\textbf{Our work}} & \textbf{Comprehensive QFL Survey} & End-to-end survey, principles, taxonomy, applications, case study, market analysis. & \textbf{N/A} & \cmark & \cmark & \cmark & \cmark & \cmark \\ \hline
        \multicolumn{9}{l}{\footnotesize{\cmark: Covered; \ \partialmark: Partially covered; \ \xmark: Not covered.}}
    \end{tabularx}
    \end{adjustbox}
\end{table*}
\subsection{Paper Structure }
\begin{figure}[htbp]
    \centering
    \includegraphics[width=0.99\linewidth]{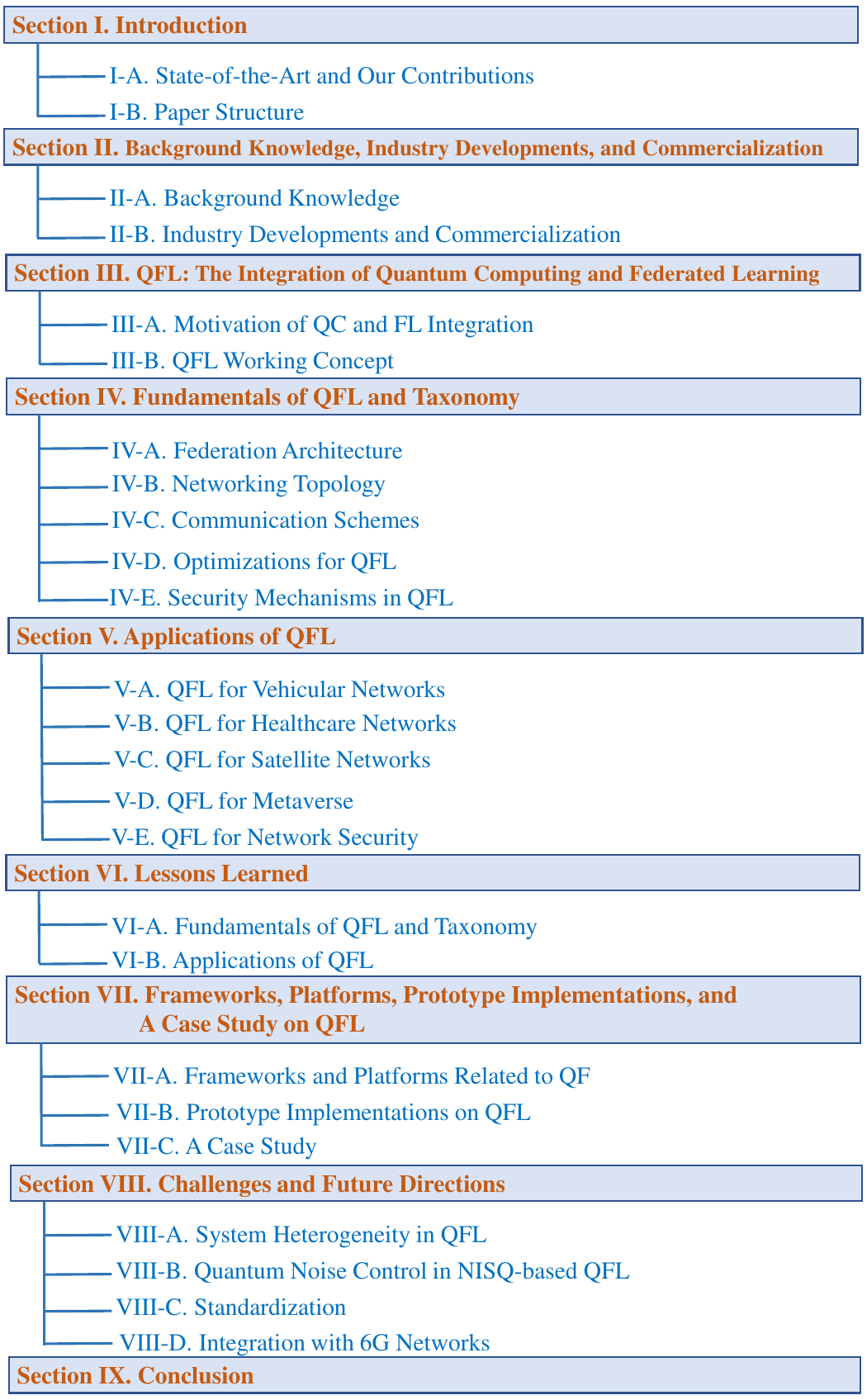}
    \caption{Organization of this survey paper.}
    \label{Fig:structure}
\end{figure}

Figure ~\ref{Fig:structure} shows how the survey is organized. Table ~\ref{tab:qfl_acronyms} has a list of the most important acronyms and abbreviations used in the paper. In Section \ref{qflbackground}, we talk about the Industry Developments and Commercialization possibilities of QFL and go over important background information about quantum computing. Read Section \ref{integrationofqcandfl} to learn more about why quantum computing and FL are being combined. Section \ref{qfltaxonomy} talks about the basics of QFL and its taxonomy. It talks about federation design, networking topology, communication schemes, ways to make QFL work better, and how to keep it safe. Section \ref{applicationsofqfl} talks about how QFL can be used in several important areas, such as healthcare networks, space networks, the metaverse, and network security. The study gave us a lot of useful information, which we have shared in the form of lessons learned on the QFL fundamentals and applications in Section \ref{lesson}. In Section \ref{casestudy}, we talk about some frameworks and platforms that are related to QFL. We also give an overview of some prototype applications on QFL and present a case study. The Section \ref{challengesandfuturedirections} lists the problems and possible solutions in the field of QFL. Finally, Section \ref{conclusions} concludes the paper.

\section{Background Knowledge, Industry Developments and Commercialization} \label{qflbackground}
This section presents a few foundational concepts related to quantum qubits, quantum gates, quantum layers, quantum measurements, and quantum model communications in a quantum network, which are essential in QFL design. Market opportunities of quantum computing and FL/ML are also discussed.

\subsection{Background Knowledge}
\begin{figure}[htbp]
    \centering
    \includegraphics[width=0.99\linewidth]{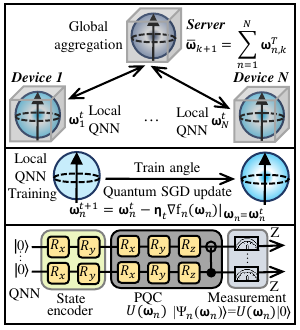}
    \caption{QFL framework where distributed quantum devices collaborate with a central server to train a shared ML model. Each device converts practical data into quantum form utilizing a state encoder, processes this data through a PQC with adjustable angle parameters, and uses the results from measurements to update local model parameters. These updated models are then sent to the server for aggregation.}
    \label{Fig:Overview}
\end{figure}

\subsubsection{Quantum Bits}
One of the fundamental units of quantum computing is quantum-bits which is also known as qubits. In traditional devices, a bit can either be in one of two states: 0 or 1. However, qubits can exist in a superposition state, meaning it can represent 0 and 1 simultaneously. In a Hilbert space, which is a two-dimensional complex vector space represented as  $\mathcal{H} \subseteq \mathbb{C}^2$, where $\mathbb{C}$ is the complex number field, while $\mathbb{C}^2$ is the space of 2D complex vectors. The fundamental computational basis states are represented as
\begin{equation}
    {|0\rangle = [1 \ 0]^T, |1\rangle = [0\ 1]^T},
\end{equation}
and these states are a complete and orthogonal set of $\mathcal{H}$ where orthogonality means \(\langle 0|1 \rangle = 0 \) and \(\langle 0|0 \rangle = \langle 1|1 \rangle = 1 \). The \textit{superposition} allows qubits to represent more than one state. A qubit state is expressed as
\begin{equation}
    |\psi\rangle = [\alpha\ \beta]^T = \alpha|0\rangle + \beta|1\rangle \in \mathcal{H},
\end{equation}
where $|\psi\rangle$ is known as \textit{ket psi} and $\alpha$ and $\beta$ are two complex numbers that must satisfy
\begin{equation}
    |\alpha|^2 + |\beta|^2 = 1.
\end{equation}

Again, the conjugate transpose of $|\psi\rangle$ is $\langle \psi|$ which is known as \textit{bra psi} \cite{chehimi2023foundations} can be written as
\begin{equation}
    \langle \psi| = |\psi\rangle^{\dagger} = [\alpha^*\ \beta^*] = \alpha^* \langle 0| + \beta^* \langle 1|.
\end{equation}

For two states $| \psi_1\rangle$ and $| \psi_2\rangle$, the inner product between two states is \begin{equation}
    \langle \psi_1|\psi_2 \rangle = \alpha_1^*\alpha_2 + \beta_1^*\beta_2,
\end{equation}
and the outer product is 
\begin{equation}
    |\psi_1\rangle \langle \psi_2| = \begin{bmatrix} \alpha_1^*\alpha_2 & \alpha_1^*\beta_2 \\ \beta_1^*\alpha_2 & \beta_1^*\beta_2 \end{bmatrix}.
\end{equation}

For a system with $D_q$ qubits, the total state space is the tensor product of the individual qubit spaces:
\begin{equation}
    \mathcal{H} = \bigotimes_{d=0}^{D-1} \mathcal{H}_d = (\mathbb{C}^2)^{\bigotimes D_q}.
\end{equation} 

\begin{figure}[htbp]
    \centering
    \includegraphics[width=0.88\linewidth]{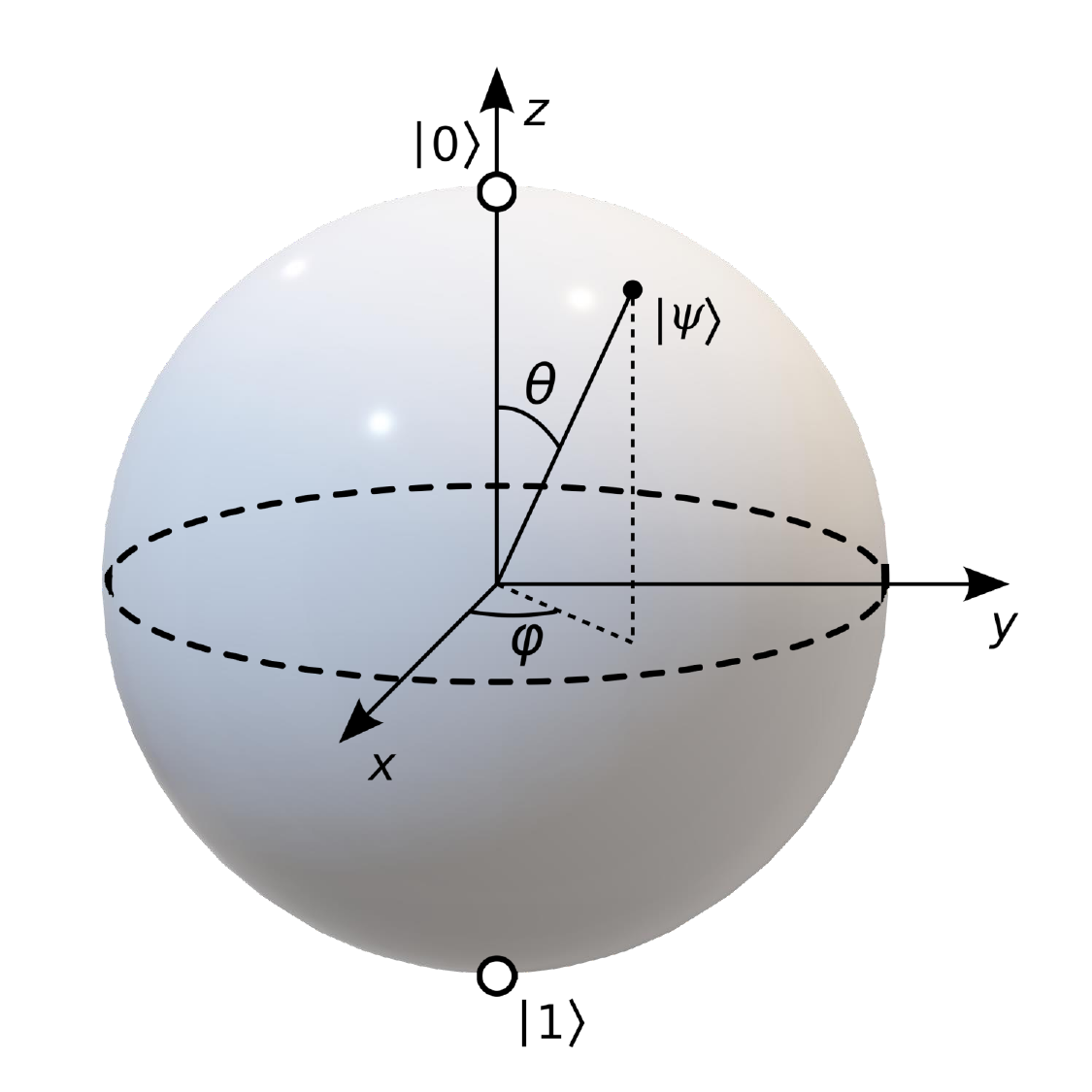}
    \caption{A Bloch sphere of a qubit where any point on the surface corresponds to a valid pure quantum state---a range of possibilities unavailable to classical bits. For instance, the state 
     \(\frac{|0\rangle + |1\rangle}{\sqrt{2}}\) lies at the equator on the positive \(x\)-axis.}
    \label{fig: qubit}
\end{figure}

In other words, given we have $D_q$ qubits, we can combine the individual spaces $\mathcal{H}_d$ into one larger space \cite{patel2025review}. A qubit can be represented using a \textit{Bloch sphere}, as shown in Fig.~\ref{fig: qubit}.

\subsubsection{Quantum Gates}
A quantum gate is defined by a unitary matrix \(U\) fulfilling \(U^\dagger U = I\) where $U^\dagger$ is the conjugate transpose of $U$ and \(I\) represents the identity matrix. This matrix converts qubit states in Hilbert space \(\mathcal{H}\), and $U$ transforms the states. The most common gates used in quantum are Pauli gates, and a \textit{single-qubit Pauli gate} can be written as
\[
X = \begin{pmatrix}
0 & 1\\
1 & 0
\end{pmatrix}, \quad
Y = \begin{pmatrix}
0 & -i\\
i & 0
\end{pmatrix}, \quad
Z = \begin{pmatrix}
1 & 0\\
0 & -1
\end{pmatrix}.
\]
Rotations with an angle \(\boldsymbol{w}\) around the \(x\), \(y\), and \(z\) axes are given as
\begin{equation}
    R_x(\boldsymbol{w}) = e^{-i \frac{\boldsymbol{w}}{2} X} = \begin{pmatrix}
\cos \frac{\boldsymbol{w}}{2} & -i \sin \frac{\boldsymbol{w}}{2} \\
-i \sin \frac{\boldsymbol{w}}{2} & \cos \frac{\boldsymbol{w}}{2}
\end{pmatrix},
\end{equation}
\begin{equation}
    R_y(\boldsymbol{w}) = e^{-i \frac{\boldsymbol{w}}{2} Y} = \begin{pmatrix}
\cos \frac{\boldsymbol{w}}{2} & -\sin \frac{\boldsymbol{w}}{2} \\
\sin \frac{\boldsymbol{w}}{2} & \cos \frac{\boldsymbol{w}}{2}
\end{pmatrix},
\end{equation}
\begin{equation}
    R_z(\boldsymbol{w}) = e^{-i \frac{\boldsymbol{w}}{2} Z} = \begin{pmatrix}
e^{-i \frac{\boldsymbol{w}}{2}} & 0 \\
0 & e^{i \frac{\boldsymbol{w}}{2}}
\end{pmatrix}.
\end{equation}

The Hadamard gate is another significant single-qubit gate widely used in QFL. It is expressed as
\begin{equation}
    H = \frac{1}{\sqrt{2}} \begin{bmatrix} 1 & 1 \\ 1 & -1 \end{bmatrix}.
\end{equation} 
Two-qubit operations can be handled as ``controlled" gates. Controlled-\(X\) (CNOT) and controlled-\(Z\) gates defined as
\begin{equation}
    C_{X,1,2} = \begin{bmatrix} I & 0 \\ 0 & X \end{bmatrix}, \end{equation} \begin{equation}
        C_{Z,1,2} = \begin{bmatrix} I & 0 \\ 0 & Z \end{bmatrix},
\end{equation}
with \(0\) being a square zero matrix and \(I\) being the identity matrix.  An explicit example of the CNOT gate for qubits labeled \((0, 1)\) is
\begin{equation}
    \text{CNOT}(0, 1) = \begin{pmatrix}
1 & 0 & 0 & 0 \\
0 & 1 & 0 & 0 \\
0 & 0 & 0 & 1 \\
0 & 0 & 1 & 0
\end{pmatrix}.
\end{equation}

Finally, the unitary operation $U(\boldsymbol{w})$ can be decomposed into simpler gates. For example
\begin{equation}
\begin{aligned}
&U(\boldsymbol{w}) = e^{i \frac{\boldsymbol{w}}{2} \sigma_0 \otimes \sigma_z} = \begin{pmatrix}
e^{i \frac{\boldsymbol{w}}{2}} & 0 & 0 & 0 \\
0 & e^{-i \frac{\boldsymbol{w}}{2}} & 0 & 0 \\
0 & 0 & e^{-i \frac{\boldsymbol{w}}{2}} & 0 \\
0 & 0 & 0 & e^{i \frac{\boldsymbol{w}}{2}}
\end{pmatrix},
\\&= \begin{pmatrix}
1 & 0 & 0 & 0 \\
0 & 1 & 0 & 0 \\
0 & 0 & 1 & 0 \\
0 & 0 & 0 & 1
\end{pmatrix}
\begin{pmatrix}
e^{i \frac{\boldsymbol{w}}{2}} & 0 & 0 & 0\\
0 & e^{-i \frac{\boldsymbol{w}}{2}} & 0 & 0\\
0 & 0 & e^{i \frac{\boldsymbol{w}}{2}} & 0\\
0 & 0 & 0 & e^{-i \frac{\boldsymbol{w}}{2}}
\end{pmatrix}
\\&\times
\begin{pmatrix}
1 & 0 & 0 & 0 \\
0 & 1 & 0 & 0 \\
0 & 0 & 1 & 0 \\
0 & 0 & 0 & 1
\end{pmatrix},
\\&= \text{CNOT}(0, 1)\{I_0 \otimes R_z(1, -\boldsymbol{w})\}\text{CNOT}(0, 1).
\end{aligned}
\end{equation}

\subsubsection{Quantum Entanglement}
Assuming we have two quantum systems $A$ and $B$ with Hilbert space $\mathcal{H}_A$ and $\mathcal{H}_B$ respectively. If we want to combine the two spaces, the total space is \(\mathcal{H}_A \otimes \mathcal{H}_B \). 

If the systems has interdependent states, for example $|\psi \rangle_A$ and $|\psi \rangle_B$, it is called \textit{separable} or unentanglement where we can simply join them as
\begin{equation}
    |\psi\rangle = |\psi\rangle_A \otimes |\psi\rangle_B.
\end{equation}

However, not all states in $\mathcal{H}_A$ and $\mathcal{H}_B$ have independent states. A general \textit{two-qubit states} can be written as
\begin{equation}
    |\psi\rangle = \sum_{a, b \in \{0,1\}} c_{ab} |a\rangle_A \otimes |b\rangle_B.
\end{equation}

We call the state \textit{entangled} if we can not factor $c_{ab}$ into a product of $c_a$ and $c_b$, consequently making $|\psi \rangle$ inseparable. For example,\textit{Bell state} ($\Omega$) is an entangled state that forms a complete basis
\begin{equation}
    \Omega = \{ |\Phi^+\rangle_{AB}, |\Phi^-\rangle_{AB}, |\Psi^+\rangle_{AB}, |\Psi^-\rangle_{AB} \}.
\end{equation}

These are defined as
\begin{subequations}
    \begin{align}
        |\Phi^+\rangle_{AB} &= \frac{1}{\sqrt{2}} (|0\rangle_A \otimes |0\rangle_B + |1\rangle_A \otimes |1\rangle_B), \label{eq:bell1} \\
        |\Phi^-\rangle_{AB} &= \frac{1}{\sqrt{2}} (|0\rangle_A \otimes |0\rangle_B - |1\rangle_A \otimes |1\rangle_B), \label{eq:bell2} \\
        |\Psi^+\rangle_{AB} &= \frac{1}{\sqrt{2}} (|0\rangle_A \otimes |1\rangle_B + |1\rangle_A \otimes |0\rangle_B), \label{eq:bell3} \\
        |\Psi^-\rangle_{AB} &= \frac{1}{\sqrt{2}} (|0\rangle_A \otimes |1\rangle_B - |1\rangle_A \otimes |0\rangle_B). \label{eq:bell4}
    \end{align}
\end{subequations}

The systems A and B are so closely connected in these states that it is impossible to classify each one as an independent state.

\begin{tcolorbox}[
    colback=blue!5!white, 
    colframe=blue!75!black, 
    title=\textbf{Tutorial Box I: Foundational Quantum Concepts},
    fonttitle=\bfseries,
    sharp corners,
    boxrule=1pt]
To grasp the innovations of QFL, a few core quantum concepts are essential. Here are some simplified, analogy-based explanations for a generalist audience.

\begin{itemize}[leftmargin=*]
    \item \textbf{Qubit:} A classical bit is like a light switch: it is definitively either OFF (0) or ON (1). A qubit, however, is like a \textbf{dimmer switch}. It can exist in a state of superposition, representing a value that is some combination of 0 and 1 simultaneously. This ability to hold more information is a fundamental source of quantum computing's power.

    \item \textbf{Superposition:} This is the state of the "dimmer switch" before you look at it. The qubit exists in a rich spectrum of possibilities. However, the moment you \textit{measure} it, the superposition "collapses," and the qubit is forced to choose a classical state—either 0 or 1. The art of quantum algorithms is to manipulate these superpositions to make the desired outcome the most probable upon measurement.

    \item \textbf{Entanglement:} Often called "spooky action at a distance," this is a unique quantum connection between two or more qubits. 

    \item \textbf{Quantum Gate:} A classical logic gate (like AND or NOT) flips bits. A quantum gate is an operation that manipulates the state of a qubit. It can be visualized as \textbf{rotating the state} of the qubit on a sphere (the Bloch sphere). These precise rotations allow for far more complex information processing than simple bit-flips.
\end{itemize}
\end{tcolorbox}

\subsubsection{Quantum Measurements}
In quantum physics, a measurement is described using an observable, which is a specific type of operator called a \textit{Hermitian operator}. A Hermitian operator \(O \) is defined as \(O = O^\dagger \), where \(O^\dagger \) represents its conjugate transpose. In projective measurements, these operators are also unitary, fulfilling \(O^\dagger O = I \). The observable may be expressed using \textit{spectral decomposition} in terms of its eigenvalues and eigenvectors
\begin{equation}
    O = \sum_{m=0}^{M-1} \lambda_m P_m,
\end{equation}
where \(\lambda_m \) are the measurement outcomes (eigenvalues), and \(P_m \) are the projectors that correspond to those results.  Each \(P_m \) projects the quantum state onto a subspace (eigenspace) associated with \(\lambda_m \). When measuring a quantum state \(|\psi\rangle \), the probability of obtaining a result \(m \) is

\begin{equation}
    p(m) = \langle \psi | P_m | \psi \rangle = \langle P_m \rangle_{\psi}.
\end{equation}

Following this measurement, the state collapses (changes) to \(P_m |\psi\rangle \), with probability $p(m)$. The expected value (average across several measurements) of the observable in state \(|\psi\rangle \) is
\begin{equation}
    E_{\psi}[O] = \sum_{m=0}^{M-1} \lambda_m p(m) = \langle O \rangle_{\psi}.
\end{equation}

Observables \(\{O_1, \dots, O_K\} \) are considered \textit{commuting} if they can be measured jointly and have similar eigenvectors. This implies that for all \(i, j \), the commutator is zero

\begin{equation}
    [O_i, O_j] = O_i O_j - O_j O_i = 0.
\end{equation}

These observables, known as \textit{pairwise commuting}, can be detected concurrently inside the same quantum system.


The choice of measuring foundation greatly determines the exact result of a quantum measurement. Different measuring bases (e.g., computational or Hadamard bases) project quantum states onto particular outcomes depending on different probability distributions. Therefore, the proper choice of measurement basis is essential to recover valuable information contained in quantum circuits \cite{patel2025quantum}. 

In quantum machine learning models, such as those applied in QFL, the measurement step usually reflects the last stage of quantum computations, so converting the quantum processing outputs into classical data usable by the next classical analysis and decision-making stages. Quantum measurement is still fundamentally probabilistic, which creates statistical uncertainty and noise problems, particularly with restricted measurements (shots) availability. Research in this crucial area is still much needed to overcome these obstacles by means of improved measurement methodologies and error reduction strategies, hence improving dependability and performance in QFL systems \cite{landi2024current}.

\subsubsection{Quantum Model Communications}
Essential in QFL, the Quantum Model Communication is the transfer of trained QML model parameters from distributed quantum devices to a central quantum server for aggregation. Following local model training with either quantum or hybrid quantum-classical algorithms, devices create model updates either in classical or quantum form, each necessitating different communication mechanisms \cite{ren2025advances}.

Standard wireless communication systems can effectively transport this data if the learned model parameters are classical, for example, numerical values generated by monitoring quantum states. Reliably enabling such transmissions is conventional communication channels such as Wi-Fi, other classical networks, mobile networks (5G and eventual 6G). Ensuring low latency and broad compatibility, these established, highly scalable, adequately resilient techniques are fit for classical parameter distribution \cite{pirandola2020advances}.

Conversely, conventional communication systems are inadequate when model parameters are encoded as quantum states. Quantum states call for specific quantum communication systems that maintain quantum coherence, such as QKD channels built using optical fiber cables or FSO networks. Through quantum entanglement and superposition, these quantum channels enable safe transmission, therefore providing intrinsically better security assurances against eavesdropping and interception than conventional techniques. Practical problems for quantum communication, however, include limited transmission rates, quantum state decoherence over long distances, and infrastructure complexity. To improve dependability and scalability in quantum model communications inside the QFL frameworks, addressing these problems calls for continuous developments in quantum repeaters, entanglement distribution techniques, and quantum networking technology \cite{pirandola2020advances}.



\subsection{Industry Developments and Commercialization}

Quantum computing, leveraging principles of quantum mechanics, promises exponential advancements in computational capabilities. With a compound annual growth rate (CAGR) of 34.8\%, Fortune Business Insights estimates that the worldwide quantum computing industry was valued at USD 885.4 million in 2023 and is expected to rise from USD 1,160.1 million in 2024 to USD 12,620.7 million by 2032 \cite{markets2027quantum}. The possibility of technology to transform sectors, including finance, healthcare, and logistics, by improved data processing and problem-solving capacity, drives this development \cite{fortune2023quantum}. 

Emphasizing even more the financial impact, a Boston Consulting Group analysis projects that by 2040, quantum computing might generate $450$ billion to $850$ billion in economic value worldwide, therefore supporting a $90$ billion to $170$ billion market for hardware and software providers. This estimate emphasizes the expected major investments and developments in the field of quantum computing \cite{bcg2024quantum}.

A subset of artificial intelligence, machine learning has evolved into a necessary tool for many different fields, fostering creativity and efficiency all around. Emerging in ML, FL allows cooperative model training across distributed devices while maintaining data privacy—a major issue in the data-driven environment of today \cite{marketus2023}.

Globally, FL is showing rather explosive expansion. Grand View Research projects the market size to be USD 119.4 million in 2022 and to rise at a CAGR of 12.7\% from 2023 to 2030 \cite{grandview2023}. Constant developments in ML methods and the growing focus on data privacy and security drive this expansion, offering a quite different viewpoint, projecting, at a CAGR of 10.6\% over the projected period, the FL market size from USD 127 million in 2023 to USD 210 million. Though forecasts vary, the increasing trend shows a strong market potential for FL systems \cite{markets2027quantum, snsinsider2023}.
There is great promise at the junction of FL and quantum computers. Aiming to maximize the computational capability of quantum computing while preserving its distributed and privacy-protecting features, QFL, integrating quantum technologies with FL, would provide hitherto unheard-of data analysis and decision-making capacity as it develops. Despite a general decline in IT spending, venture capital financing in quantum computing is rising to \$1.2 billion in 2023. This financial dedication underscores the rising hope in quantum technologies' ability to create a major economic impact \cite{businessinsider2025quantum}.



\section {QFL: The Integration of Quantum Computing and Federated Learning} \label{integrationofqcandfl}

\subsection{Motivation of QC and FL Integration}
To highlight the motivation behind integrating QC and FL, it is important to consider the unique strengths and properties of each technology and how their combination can create a powerful new paradigm. 

On one side, QC offers exceptional computational capabilities that go far beyond what is possible with classical algorithms. By leveraging quantum principles such as superposition and entanglement, QC enables novel algorithms capable of solving complex problems intractable for classical systems \cite{yazdi2024application}. For example, in QML algorithms such as the VQE have shown the potential to outperform classical machine learning models, especially in areas such as optimization and simulation. 

\begin{itemize}
    \item \textit{Solving High-Dimensional Optimization Problems Efficiently:} Classical FL systems often struggle with complex optimization tasks, especially when dealing with large-scale, non-convex loss surfaces typical in deep learning. QC introduces quantum optimization algorithms, such as the VQE and Quantum Approximate Optimization Algorithm (QAOA), which leverage quantum parallelism to explore vast solution spaces more efficiently. These quantum methods can potentially find better minima in high-dimensional landscapes faster than classical optimizers, leading to improved convergence and accuracy in FL tasks \cite{adediran2024advancements}.
     \item \textit{ Processing Quantum Data and Modeling Quantum-Influenced Systems:} Classical FL cannot handle inherently quantum data or systems influenced by quantum mechanics, such as in quantum chemistry, materials science, or quantum sensor networks. QC is naturally suited to represent and manipulate quantum states, enabling the training of QML models directly on quantum information \cite{zaman2023survey}. When integrated into FL systems, QC expands the applicability of FL to quantum-domain tasks that are beyond the reach of classical data representations or models.
  \item \textit{Reducing Communication Bottlenecks via Quantum Encoding and Compression:} One of the main limitations in classical FL is communication overhead due to frequent transmission of large model updates. QC offers novel quantum communication techniques, such as quantum teleportation and entanglement-assisted compression, which can drastically reduce the volume of information shared across nodes \cite{li2023entanglement}. These techniques enable more efficient aggregation of quantum model updates in distributed settings, improving scalability and responsiveness in large-scale federated networks.
\end{itemize}

On the other hand, FL provides a decentralized approach to training machine learning models while preserving data privacy. Instead of aggregating raw data in a central location, FL allows data to remain local, with only model updates shared among participants. This approach enhances privacy, reduces communication costs, and enables learning across distributed and siloed data sources. As a result, FL has become especially valuable in privacy-sensitive quantum application domains such as IoTs and wireless networks \cite{fouda2024privacy}.

\begin{itemize}
    \item \textit{Preserving Data Privacy in Quantum-Enhanced Applications: } QML models that collect and aggregate quantum or classical data into a central processor or server pose privacy and security risks, particularly in sensitive fields like healthcare, finance, and personal devices \cite{tomar2025comprehensive}. FL keeps quantum-enhanced data confined to quantum nodes or organizations. FL ensures data privacy while facilitating collaborative learning by exchanging encrypted or abstracted model changes rather than raw data. In quantum-enabled systems, this is essential for user trust and HIPAA and GDPR compliance.

    \item \textit{Enabling Scalable and Collaborative Quantum Learning Across Distributed Devices:} As QC evolves, the vision of interconnected, modular quantum processors becomes more feasible. FL supports this model by enabling decentralized learning across multiple quantum nodes, each operating with limited qubits and computational power. Unlike centralized QML, which depends on a single large-scale quantum device, FL allows many smaller, geographically dispersed quantum systems to collaboratively train a global model. This decentralization aligns perfectly with the distributed architecture of quantum networks and helps overcome current NISQ-era hardware constraints \cite{khalid2023quantum}.

    \item \textit{Reducing Communication and Resource Bottlenecks in Quantum Networks:} Centralized QML often faces scalability challenges due to the need to transmit large volumes of quantum data or intermediate states to a central processor, which is impractical over current quantum communication infrastructure \cite{rishiwal2025new}. FL mitigates this issue by minimizing communication requirements: only quantum model updates or compressed parameters need to be shared among participants. This not only reduces communication overhead but also preserves bandwidth and hardware resources, making intelligent quantum systems more efficient and practical to deploy across edge quantum devices or hybrid quantum-classical platforms.
\end{itemize}

After evaluating the field's substantial state-of-the-art research, merging QC and FL is motivated by their complimentary strengths and limits. QC addresses computational intensity and scalability issues in FL, whereas FL provides privacy-preserving and decentralized framework for practical implementation in distributed quantum settings \cite{xu2022privacy}. It constitute QFL, a promising new paradigm that incorporates both technologies' capabilities.

QC and FL integration in QFL provides a solution to difficulties in both domains, creating new prospects for safe, intelligent quantum learning systems \cite{kang2024quantum}. QFL might power a wide range of real-world applications and shape quantum-driven technologies by accelerating scientific discovery and improving data privacy in quantum-enabled infrastructures.

\subsection{QFL Working Concept}
In this section, we present the architecture and fundamental operation of QFL, as illustrated in Fig.~\ref{Fig:Overview}. QFL represents a novel paradigm that integrates QC with FL principles, allowing multiple quantum-enabled devices to collaboratively train a shared machine learning model without exchanging raw data \cite{dai2025state}. This decentralized approach not only enhances data privacy but also leverages the computational advantages of quantum devices in distributed environments. The overall QFL process is coordinated by a central server, which manages communication, aggregation, and synchronization between quantum devices \cite{sahu2024nac, rahman2025sporadic}. A typical QFL algorithm follows a cyclical, multi-step process composed of the following key stages:

\begin{itemize}
    \item \textit{Step 1 – Quantum Data Encoding: } 
    Each device begins by encoding its private classical dataset into quantum states. This is performed using a quantum state encoder, which maps classical feature vectors into quantum Hilbert space representations \cite{hur2022quantum}. The encoding process generally starts with state preparation, where qubits are initialized in their ground state, $|0\rangle$. Various encoding techniques—such as angle encoding, amplitude encoding, or basis encoding—may be employed depending on the nature of the data and the desired learning task. This step ensures that the data is represented in a form compatible with quantum processing and suitable for input into PQCs \cite{khan2024beyond}.

    \item \textit{Step 2 – Local Training with PQCs:}
    Once the data is encoded, it is passed through a PQC at each device. The PQC consists of a series of quantum gates whose behavior is controlled by trainable parameters (analogous to weights in classical neural networks) \cite{du2021learnability}. These gates manipulate the qubits and entangle them to explore complex quantum correlations within the input data. After applying the PQC, the qubits are measured, and the resulting classical outputs (e.g., measurement probabilities or expectation values) are used to compute the loss function. Using gradient-based optimization techniques such as Stochastic Gradient Descent (SGD), the device updates its local PQC parameters based on the loss, effectively training the quantum model on local data \cite{wang2022qoc}.

    \item \textit{Step 3 – Model Sharing and Aggregation:} Every quantum device provides its updated PQC parameters or gradients to a central server \cite{luo2018parameter} following local training. Usually, applying techniques such as FedAvg, the server generates an updated global model by means of a safe and privacy-preserving aggregate. After that, this model is sent back to every involved quantum device to synchronize their local PQC settings. One worldwide cycle is completed here; the procedure is continued over several rounds until the model converges or achieves acceptable performance. Although this procedure is comparable to classical FL, QFL adds special factors including quantum encoding, circuit depth restrictions, and quantum-specific noise management. These variations make QFL more appropriate for quantum data processing and privacy-sensitive uses in quantum-enabled environments. This difference clarifies for researchers familiar with classical FL the motivation and required modifications for working in QFL environments \cite{paul2022towards}.
\end{itemize}

Throughout the QFL process, no raw data ever leaves the local device environments, ensuring strong data privacy and compliance with regulatory requirements. Moreover, the use of quantum circuits in the training process introduces potential computational advantages, especially for learning tasks involving high-dimensional data or complex feature interactions.

\color{black}

\begin{figure*}[htbp]
    \centering
    \includegraphics[width=0.99\linewidth]{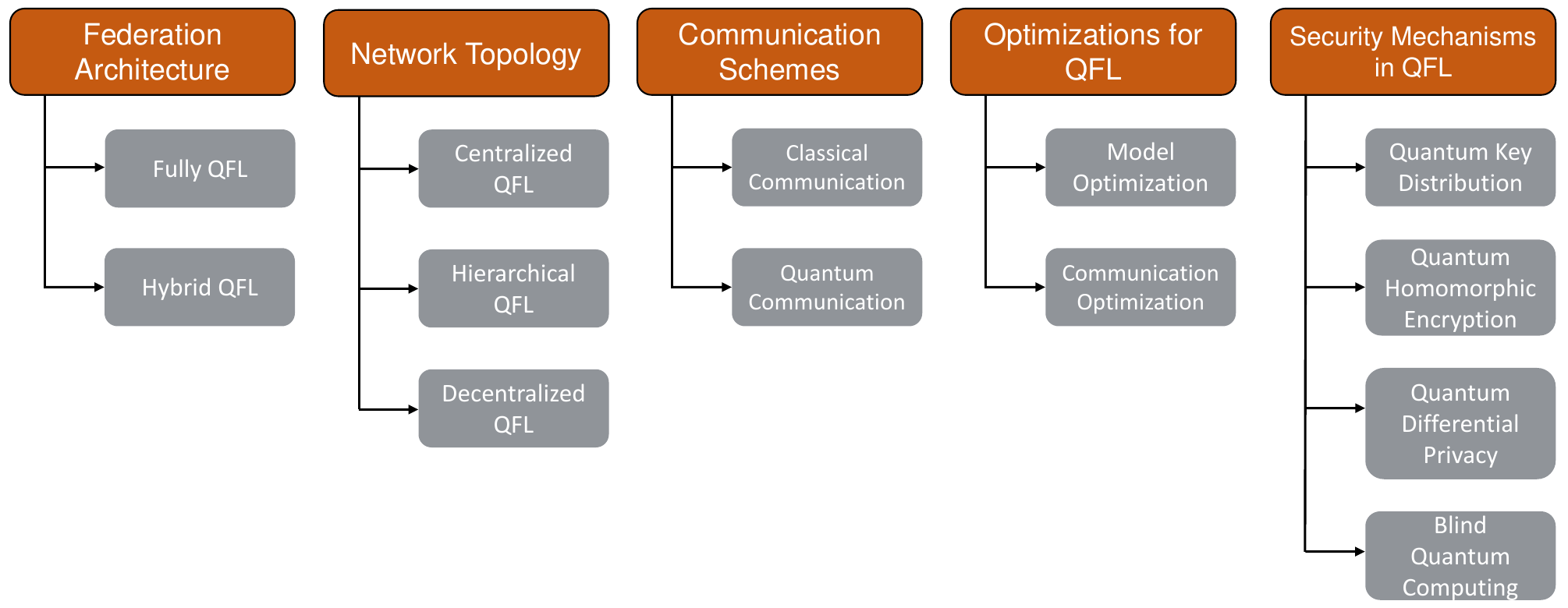}
    \caption{QFL fundamentals explored in this paper. }
    \label{Fig:fundamentals}
\end{figure*}

\section {Fundamentals of QFL and Taxonomy} \label{qfltaxonomy}
\color{black}
Having established the motivation for integrating quantum computing with federated learning, this section delves into the core components that constitute a QFL system. For a communications generalist, this section provides a framework for building a QFL network. We present a holistic discussion on the fundamentals (Fig. ~\ref{Fig:fundamentals}), including Federation Architecture, Networking Topology, Communication Schemes, Optimization in QFL, and Security Mechanisms in QFL. We also provide a taxonomy summarized in Table ~\ref{tab:qfl_fundamentals_cos}, giving valuable insights into these design aspects of the QFL system. 
\color{black}

\subsection {Federation Architecture}
\subsubsection {Fully QFL}
In Fully Quantum Federated Learning (Fully QFL), numerous dispersed devices with fully quantum models cooperate safely and effectively to train a global quantum model without explicitly communicating sensitive local quantum data.  A common example is when all devices use VQE with PQCs, enabling improved computations and optimization at local quantum nodes \cite{huang2022quantum}.

 This design uses a hybrid quantum-classical technique to train a local quantum model on each device, namely a PQC with classically optimal parameters to lower a preset quantum loss function.  PQCs, typically quantum gates with changeable classical parameters, enable quantum hardware to learn and embed complex patterns in quantum states \cite{abou2024privacy}.  Customers securely upload quantum model updates or measurement results to a quantum-enabled server after local training.  This server creates a global quantum model from quantum updates using properly prepared quantum aggregation or quantum averaging and sends updated parameters to devices.  This method utilizes quantum entanglement, superposition, and interference to code and analyze data more efficiently than traditional federated methods \cite{park2025entanglement}.

 Quantum features prevent undesired inference of device data from aggregated quantum states, making fully QFL designs more secure and private.  Adversarial attacks, unauthorized data breaches, and FL system faults are reduced by this architecture.  Recent research suggests that fully quantum federated architectures can transform privacy-preserving distributed quantum machine learning, leading to widespread adoption in decentralized scenarios \cite{innan2024fedqnn}, despite challenges such as quantum noise, hardware constraints, and maintaining quantum coherence.

\subsubsection {Hybrid QFL}


Combining classical NN layers with quantum layers, the Hybrid QFL architecture essentially merges classical machine learning capabilities with quantum-enhanced computations to use the benefits of both paradigms \cite{hisamori2024hybrid}. Usually consisting of convolutional or fully connected layers that effectively handle large-dimensional data and the lower computational overhead before quantum processing, initial data processing and feature extraction usually occur in such hybrid architectures via classical NN layers. Extracted classical features are then encoded into quantum states and run through PQCs, therefore allowing complicated pattern recognition, enhanced feature mapping, and optimization powers given by quantum mechanics \cite{abbas2021power}.

A typical hybrid QFL design consists of two initial conventional NN layers that are used for feature extraction and dimensionality reduction. Subsequently, quantum layers are employed to encode and learn high-level features using quantum gates and variational quantum circuits. These quantum layers consist of quantum circuits parameterized by trainable classical parameters, refined using hybrid quantum-classical training methods like VQAs. By means of classical optimization approaches, VQAs enable quantum circuits to iteratively learn optimal parameters, hence greatly enhancing learning performance over classical-only models in several challenging tasks \cite{jaksch2023variational}.

Multiple distributed devices independently learn their local hybrid quantum-classical models on local datasets in hybrid QFL. Usually, by averaging classical parameters, the devices then safely send classical parameters of quantum layers or measurement results to a centralized aggregator, which accumulates the updates. Aggregated parameters are then provided to all devices, therefore improving global model performance cooperatively without specifically disclosing sensitive data \cite{ren2023qfdsa}. By using quantum properties—such as quantum uncertainty and entanglement—to further safeguard data privacy, hence greatly lowering the risks connected with adversarial attacks, privacy breaches, or illegal data inference, hybrid QFL designs offer excellent security guarantees.

\subsection {Networking Topology}
\subsubsection {Centralized QFL}
In Centralized QFL, a central quantum \textcolor{black}{server organizes} communications and model aggregation across several distributed quantum devices using a hub-and-spoke model of the networking topology. During the FL process, the networking topology guarantees scalability and manageability across quantum nodes by allowing organized coordination and effective parameter synchronizing \cite{qiao2024transitioning}. Equipped with either local quantum models or hybrid quantum-classical models, every device trains on its private data and sends encrypted updates, such as quantum measurement results or PQC parameters, to the central aggregator. The central server then distributes the revised global model back to all devices by combining these changes with quantum-aware aggregation methods.

Among the various benefits of the centralized topology in QFL are simplified communication protocols, reduced system complexity, and better control over synchronizing and training convergence. To guarantee safe communication and guard against eavesdropping or hostile interference \cite{singh2021quantum}, the central server can also apply sophisticated quantum cryptography techniques such as QKD or BQC. This design does, however, also bring a possible single point of failure; therefore, the central server becomes an important target for quantum-level attacks or performance constraints.

Practically, Centralized QFL utilizes quantum communication channels, including quantum repeaters or entangled photon networks, for the secure and efficient transmission of quantum information over extensive distances. Maintaining coherence and integrity of quantum states during transmission depends on these quantum communication systems, hence guaranteeing the dependability of FL systems \cite{chehimi2023foundations}. Especially in settings where a resourceful central server can be trusted and maintained, centralized QFL topologies remain a basic concept for establishing secure and privacy-preserving quantum machine learning systems at scale despite technical constraints.

\subsubsection {Hierarchical QFL}
\begin{figure*}[htbp]
    \centering
    \includegraphics[width=0.99\linewidth]{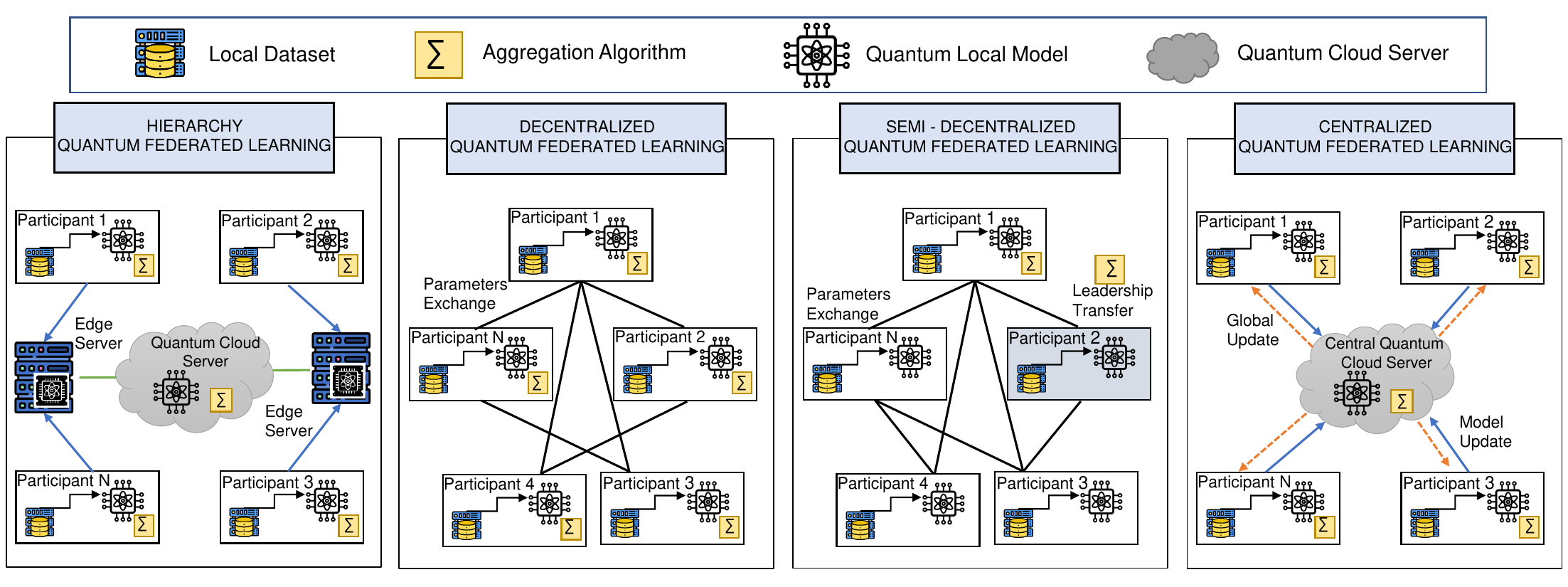}
    \caption{Illustration of QFL Architectural Paradigms Based on Decentralization Levels: A Comparative Overview of Hierarchical, Decentralized, Semi-Decentralized, and Centralized QFL Approaches Including Communication Schemes, Aggregation Methods, and Quantum Model Distribution}
    \label{Fig:decentralization}
\end{figure*}

A multi-tiered networking design is presented in Hierarchical QFL to improve scalability, lower latency, and control quantum resource distribution over dispersed systems of vast scale as displayed in Fig. ~\ref{Fig:decentralization}. Though hierarchical models have been extensively investigated in classical FL, such as device-edge-cloud systems \cite{su2025joint}, these ideas can be extended and adapted for quantum settings to enable more efficient quantum learning frameworks. Under this design, quantum edge servers connect with a central quantum cloud server at the upper layer while quantum devices and quantum edge servers constitute the lower tier. While protecting quantum data privacy, this tiered communication hierarchy enhances training efficiency and system scalability.

With PQC or VQA, every quantum device trains a local quantum or hybrid quantum-classical model. The relevant quantum edge server receives the locally trained parameters or measurement statistics from an intermediary aggregator. Like classical edge nodes in hierarchical FL, these edge servers do intra-cluster aggregation and then transfer the condensed model parameters to a central quantum cloud server. This two-stage aggregation lowers communication overhead by load balancing among quantum resources.

Inspired by classical systems such as Hierarchical FL across Heterogeneous Cellular Networks \cite{abad2020hierarchical} and Dynamic Resource Allocation in Decentralized Edge Intelligence \cite{lim2021decentralized}, hierarchical QFL could dynamically allocate quantum communication and computation resources between levels. It also allows fault-tolerant and adaptive learning, in which case quantum edge servers can replace local coordination without depending just on a central node. Maintaining fidelity over hierarchical links depends on quantum communication methods including entanglement switching or quantum repeaters.

\subsubsection {Decentralized QFL}
In Decentralized QFL, several quantum devices cooperatively train a shared global quantum model without depending on a central aggregator by use of a P2P network built as such. Inspired by classical decentralized FL systems, such as those described in the distributed edge intelligence frameworks, each quantum device independently trains its quantum or hybrid quantum-classical model, usually using PQCs and VQA \cite{kannan2024quantum}. Following local training, every device directly exchanges quantum model parameters or quantum measurement results with surrounding devices over safe quantum communication channels, such as entanglement-assisted quantum networks or QKD links.

Distributed QFL removes single points of failure, unlike centralized topologies, therefore enhancing fault tolerance, scalability, and resistance against adversarial assaults. Using distributed protocols such as quantum consensus algorithms, quantum-secured gossip protocols, or blockchain-inspired synchronizing mechanisms, quantum devices communicate following distributed protocols, hence enabling collective convergence toward a global model. Because only encrypted or aggregated quantum model parameters are communicated and raw data stays localized, this distributed structure also improves data privacy \cite{malina2021post}.

Entanglement swapping and quantum teleportation, among other quantum communication methods, enable safe, direct quantum connections between nodes, hence enhancing security guarantees and preserving the quantum coherence required for distributed quantum computations \cite{ghaderibaneh2022efficient}. Decentralized QFL naturally adapts to changing network topologies and device participation, hence, it is especially appropriate for quantum-enhanced edge computing environments, quantum IoT applications, and scenarios with constrained infrastructure.

\color{black}
When comparing these networking topologies, a clear trade-off emerges between simplicity, scalability, and robustness. The \textbf{Centralized} approach is the most straightforward to implement and manage but is inherently limited by the server's capacity and represents a single point of failure, making it preferable for smaller-scale or experimental QFL systems where control is paramount \cite{quy2024federated}. \textbf{Hierarchical QFL} offers a more balanced solution, improving scalability and reducing communication bottlenecks, which is highly effective in contexts like edge computing where a natural tiered structure exists \cite{wang2023quantum}. However, it introduces additional complexity in managing intermediate aggregators. \textbf{Decentralized QFL} provides the highest level of fault tolerance and privacy, making it theoretically ideal for large-scale, trustless environments \cite{gurung2023decentralized}. However, it can introduce significant communication overhead due to increased peer-to-peer coordination. Therefore, its preference is limited by significant practical challenges in achieving network-wide consensus and synchronization, which remains an active area of research.
\color{black}

\subsection {Communication Schemes}
\subsubsection {Classical Communications}
A quantum-empowered FL system, especially intended for Space-Air-Ground Integrated Networks (SAGIN), which are fundamental for the evolution of future 6G networks. It essentially manages the complexity of large datasets and model training over conventional communications networks like 5G/6G by using variational quantum algorithms for local training and quantum relays for safe data delivery. All things considered, the architecture greatly improves SAGIN security and efficiency of data processing, providing a strong means of developing edge intelligence applications in the era of 6G technology \cite{wang2023quantum}.

The authors \cite{park2024dynamic} present a new QFL framework for satellite-ground communication using low Earth Orbit (LEO) satellite capability to improve world connectivity. Using Scalable Quantum Neural Networks (sQNN), which adapt to various configurations—angle and pole—the framework, called SQFL, optimizes both computing and communication workloads. Moreover, the efficiency of data transmission across conventional communications networks such as 5G/6G is much improved by including superposition coding and consecutive decoding techniques. Under low signal-to-noise ratios and non-IID data, especially, extensive experimental results show that SQFL provides better performance in computational speed and communication efficiency than conventional FL methods in multi-LEO satellite scenarios and optimizing power allocation for next-generation 6G networks.

Moreover, artificial intelligence is discussed especially in the development of 6G networks. Using FL and QFL, the work concentrates on satisfying the requirements for computational efficiency and privacy in the heterogeneous and distributed SAGIN configuration. It underlines numerous uses where FL and QFL improve the functionality of SAGINs, including a thorough case study on QFL inside UAV networks, therefore proving its benefits over conventional FL techniques in terms of training efficiency and data privacy. Furthermore, underlining its potential to greatly expand real-time intelligent applications across several fields, the authors emphasize important research difficulties and the need to standardize QFL to enable its widespread use in future 6G networks \cite{quy2024federated}.

\subsubsection {Quantum Communications}
\color{black}
Quantum teleportation and entanglement-assisted communication are two examples of quantum communication protocols that have a lot of potential in theory. However, their use in QFL systems is limited by the present NISQ hardware. The short time that qubits can stay coherent is a big problem. Interactions with their surroundings make qubits lose their quantum state quickly, a process called decoherence. This brief period of stability, which is usually measured in microseconds, makes it very hard to undertake quantum computations and communications that last a long time or are very complicated. This means that for QFL, quantum models need to be sent and processed before the qubits decohere. This is a challenging problem for distributed systems that function over long distances or with high network delay \cite{ahmed2025osi}.

Two other important hardware problems are low gate fidelity and the difficulty of distributing entanglement. Quantum gates, which are the basic parts of quantum algorithms, include built-in error rates. Low gate fidelity indicates that when operations are done on qubits, they can make mistakes that change the quantum state, which can lead to wrong results. Likewise, distributing high-quality entangled qubit pairs over long distances is necessary for protocols like teleportation, but it's very hard to do because of channel noise and decoherence, which leads to low entanglement fidelity and slow distribution rates. These limitations from the NISQ paradigm make it very hard to communicate using QFL, and any practical QFL communication scheme must take these hardware flaws into account. This often means using advanced quantum error correction methods that are still in the early stages of development \cite{chen2024nisq}.
\color{black}

A sophisticated optical quantum communication system designed for free space situations was improved by optical combining methods and a generic Kennedy receiver. Based on a thorough quantum channel model generated using the P-representation, the system uses a new conditional dynamics-based Kennedy receiver with threshold detection to overcome turbulence and thermal noise. In situations with low thermal noise or weak turbulence, the suggested system not only minimizes environmental interferences efficiently but also surpasses the traditional quantum homodyne receiver limit, thus establishing a new benchmark in quantum communication performance \cite{yuan2020free}. On the other hand, two creative power allocation techniques for multi-hop FSO networks intended to maximize system performance are presented to minimize upper limits on error probability under quantum limit conditions \cite{pokharel2025quantum}. The efficacy of the initial method is assessed by the average error probability that is derived through numerical analysis. The second strategy, which employs closed-form error expressions, simplifies the system at the expense of some performance degradation. By lowering fading effects and adjusting the power allocation based on the complete information of the channel state information (CSI), these techniques greatly improve the dependability and efficiency of the FSO networks overall, exceeding conventional equal power distribution techniques in preserving signal integrity over long distances \cite{abou2012power}.

Lastly, the evolution and performance of relay-assisted satellite FSO systems employing quantum key distribution for improving vehicle network security are investigated in this article. High-altitude platforms are used as relays to handle issues such as transceiver misalignment, velocity fluctuations, noise, and atmospheric turbulence, therefore influencing quantum bit error rates and secret-key generation rates. Extensive Monte-Carlo simulations under well-selected system parameters—such as intensity modulation depth and amplifier gain—show clearly how much safe communication efficiency in vehicle networks may be improved \cite{vu2020design}.

\subsection {Optimizations for QFL}

\subsubsection {Model Optimization}
Within the context of QFL, model optimization plays a crucial role, which refers to the process of refining quantum algorithms to enhance the performance of quantum models while optimizing resource usage and minimizing computational overhead. Model optimization in QFL is pivotal for several reasons and involves a series of techniques and strategies designed to improve the efficiency of QNNs deployed across multiple decentralized nodes, i.e., quantum devices. Quantum devices operate under significant physical and technological constraints, for example limited number of qubits, error-prone operations, and short coherence times. Optimizing QFL models enables proper use of the available quantum resources by reducing circuit depth and the number of required qubits. Besides, model optimization for QFL also reduces the impact of errors on overall model performance. These are highly sensitive quantum systems. Moreover, in this era of ever-growing networks, it is important to ensure that a QFL algorithm can be extended to larger networks with numerous quantum nodes without increasing the computational overhead. Effective model optimization strategies enable QFL to scale while maintaining or improving the speed and accuracy of quantum computations. Besides, QFL enhances data privacy as data remains local to each node. Enhancing data encoding and quantum state preparation through different model optimization techniques can lead to more robust models immune to potential quantum threats and eavesdropping, thereby increasing the overall security of the network. 

Several optimization techniques are available for enhancing the performance of QFL models. Quantum natural gradient descent (QNGD) integrates the natural gradient descent concept with QC \cite{qi2024federated}. Traditional gradient descent approaches assume a Euclidean geometry where all directions in the parameter space are treated uniformly. However, in the quantum world, this assumption does not hold because the geometry related to quantum parameter spaces is inherently non-Euclidean. This means that the shortest path between two points in the parameter space of quantum states follows a complex trajectory rather than being a straight line. Overlooking this curvature may result in suboptimal updates that fail to accurately represent the actual landscape of the model's parameters. Furthermore, traditional gradient descent algorithms often exhibit slower convergence rates or may become trapped in local minima, leading to inefficient training paths. QNGD addresses this by tackling this issue by adjusting the update steps according to the quantum Fisher information metric. This results in a more efficient optimization path, leading to faster convergence and potentially lower quantum resource usage \cite{qi2023optimizing}.

Besides, the VQE is a hybrid quantum-classical algorithm that has been effectively adapted for use within QFL. It was initially developed for quantum chemistry applications to find the ground state energies of molecular systems. Minimizing a loss function is analogous to finding the ground state of a Hamiltonian in VQE. This approach not only enhances computational efficiency and accuracy but also maintains the decentralized and privacy-preserving nature of QFL. Thus, VQE serves as a robust method for leveraging quantum mechanics in distributed quantum learning networks \cite{liu2022layer}.

Individual models can also be optimized at each node in a QFL framework using quantum annealing (QA). QA finds the optimal parameters for the quantum circuits or models used at each node by minimizing a quantum Hamiltonian that represents the loss function of the QFL model. More specifically, this approach is particularly designed to travel across complex energy landscapes by gradually transitioning from a quantum superposition of many possible states towards the state that minimizes the overall system's energy, which represents the optimal solution. Thus, QA is well-known for its ability to escape local minima, which is a common issue faced while solving complex optimization problems \cite{yulianti2022implementation}. All these optimization techniques are quantum-based and can perform optimization tasks faster than classical algorithms. Especially, problems where the solution space grows exponentially with the number of variables can be solved more efficiently using quantum-centric approaches. Moreover, as the number of nodes in QFL increases, so does the complexity of the optimization problem. \textcolor{black}{These quantum-based optimization techniques scale well in such scenarios, handling complex and high-dimensional landscapes more effectively than classical approaches \cite{ansere2025quantum}}.

\color{black}
When comparing these optimization techniques, it becomes clear that no single method is universally superior; instead, their effectiveness is highly context-dependent. For instance, \textbf{QA} is highly effective for specific combinatorial optimization problems that can be mapped to an Ising model, leveraging specialized hardware to find global minima \cite{bozejko2024optimal}. Its application, however, is narrow and not suited for general-purpose QML model training. In contrast, \textbf{VQE} and \textbf{QNGD} are more versatile as they are designed for training variational quantum circuits. The trade-off lies in their complexity and resource requirements. \textbf{VQE} is more experimentally mature and simpler to implement on near-term NISQ devices, but it can be susceptible to barren plateaus and require an enormous number of measurements  \cite{colella2025variational}. \textbf{QNGD}, while theoretically more powerful due to its ability to navigate the complex geometry of the quantum parameter space for faster convergence, carries a significant computational overhead in calculating the quantum Fisher information metric, making it more challenging to scale on current noisy hardware \cite{sohail2025quantum}. The choice of optimizer is therefore a critical trade-off: QA is preferable for specialized optimization tasks, VQE is the pragmatic choice for near-term variational circuit training, and QNGD represents a more resource-intensive but potentially more efficient path for future, more capable quantum processors.
\color{black}

\color{black}
Although there is a lot of potential in these optimization strategies, their practical maturity differs. Numerous methods have been investigated in simulations; nevertheless, experimental verification on actual quantum hardware is still a significant obstacle, frequently restricted to proof-of-concept demonstrations on a tiny scale. Table ~\ref{tab:optimization_summary_grid} lists the main QFL model optimization methods, together with their validation status, common uses, and main limitations. This comparison provides a more realistic view of their current preparedness for practical QFL implementation by helping to differentiate between approaches that are primarily theoretical and those that have demonstrated feasibility in experimental settings.

\begin{table*}[h!]
    \color{black}
    \centering
    \caption{Summary of QFL Model Optimization Techniques.}
    \label{tab:optimization_summary_grid}
    \small
    \renewcommand{\arraystretch}{1.8} 
    \begin{tabularx}{\textwidth}{|l|X|X|X|}
        \hline
        \textbf{Technique} & \textbf{Validation Status} & \textbf{Primary Application in QFL} & \textbf{Key Constraints} \\ 
        \hline
        
        \textbf{Quantum Natural Gradient Descent (QNGD) \cite{sohail2025quantum, minervini2025quantum}} & 
        Primarily \textbf{Simulated}; some small-scale experimental demonstrations exist. & 
        Training variational quantum circuits (e.g., QNNs) by respecting the non-Euclidean geometry of quantum state space. & 
        High computational cost to calculate the quantum Fisher information metric; highly sensitive to hardware noise. \\ 
        \hline
        
        \textbf{Variational Quantum Eigensolver (VQE) \cite{colella2025variational, sobhani2025variational}} & 
        \textbf{Experimentally Validated} on various NISQ devices for small-scale problems like molecular simulations. & 
        Finding optimal model parameters by mapping the loss function to a solvable Hamiltonian, which is then minimized. & 
        Can suffer from ``barren plateaus'' (vanishing gradients) in deep circuits; requires a large number of circuit measurements. \\ 
        \hline
        
        \textbf{Quantum Annealing (QA) \cite{yulianti2022implementation, bozejko2024optimal}} & 
        \textbf{Experimentally Validated} on commercial quantum annealers (e.g., D-Wave systems). & 
        Solving complex combinatorial optimization problems that can be mapped to an Ising model Hamiltonian. & 
        Limited to optimization problems and is not a universal model of quantum computation; hardware has limited qubit connectivity. \\ 
        
        \hline
    \end{tabularx}
\end{table*}
\color{black}

\subsubsection {Communication Optimization}
While model optimization in QFL involves refining the QML algorithms themselves to improve training speed, accuracy, and resource utilization, communication optimization in QFL concentrates specifically on enhancing the efficiency and security of data transfer within QFL systems \cite{xu2022privacy}. Techniques such as quantum teleportation or entanglement-based protocols are used to reduce latency and energy consumption by optimizing resource allocation, channel selection, and bandwidth across devices and infrastructure \cite{chehimi2025reconfigurable}. We note that minimizing latency is crucial for synchronizing updates across distributed quantum nodes to facilitate timely aggregation in the server within a QFL framework, thereby enhancing the overall speed and responsiveness of the learning process. By selecting optimal communication channels and efficiently, allocating bandwidth efficiently, the system can handle large volumes of data with minimal delay, thereby enhancing the overall network performance. For instance, a communication-efficient QFL framework can be adapted in the healthcare industry to ensure patient privacy while optimizing data transmission rates and processing power across quantum networks. This helps reduce operational costs and accelerates the timelines for diagnostics and treatments, showcasing the transformative potential of QFL in healthcare \cite{bhatia2024communication}. 

Moreover, quantum devices currently face significant resource constraints, including limited qubit availability and short coherence times. By optimizing resource allocation, energy consumption is lowered, making the system more sustainable and cost-effective. Quantum-enhanced communication protocols help in maintaining the integrity and security of data transmission, which is pivotal given the sensitive nature of the data often handled in QFL setups. 

Incorporating techniques, i.e., quantum key distribution (QKD), quantum teleportation, entanglement rate, and QAOA, can significantly enhance the resource allocation, channel selection, and bandwidth optimization strategies for communication efficiency in QFL environments. QKD is a secure communication method to enable two parties to generate a shared random secret key known only to them that can be used to encrypt and decrypt messages. It is considered highly secure because it is based on the laws of quantum mechanics, rather than on mathematical problems that could potentially be solved with sufficient computing power. Hence, QKD can be used to secure the communication channels in the QFL network \cite{kaewpuang2023adaptive}, ensuring that the transmission of model updates between nodes is protected against eavesdropping. QKD resources can be allocated in a dynamic fashion based on current network security needs so that the system can maintain high security without excessive usage of quantum resources like entangled photons, which are also needed for other tasks like quantum teleportation.

Quantum teleportation is a quantum-enhanced communication technique by which quantum information (such as quantum states representing model parameters) can be transmitted from one location to another. It harnesses the synergy of classical communication and quantum entanglement to facilitate the transmission process. Quantum teleportation can be employed to transmit quantum information directly between nodes without having to send the physical qubits through potentially insecure channels. Since the quantum state itself carries the necessary information, this method can optimize bandwidth usage by reducing the amount of classical information needed for model updates. Moreover, the rate of quantum teleportation can be adjusted based on the bandwidth availability and the urgency of the updates, helping to manage network load and prioritize tasks \cite{narottama2023federated}. Another quantum-centric optimization algorithm is QAOA, which offers a promising approach for optimizing communication in QFL environments. It leverages the principles of quantum superposition and entanglement to find the optimal configuration of network parameters that minimize latency, energy consumption, and maximize bandwidth utilization across a distributed quantum network. These optimization techniques are specifically crucial in QFL, where efficient communication is important for synchronizing model updates without excessive overhead \cite{choi2019tutorial}.

\subsection {Security Mechanisms in QFL}
\subsubsection {Quantum Key Distribution}
QFL improves security and privacy by integrating QC and FL systems utilizing the concepts of QKD. QKD is a quantum communication system that enables two distant participants to generate shared random cryptographic keys whose secrecy is guaranteed securely. It is inspired by the fundamental laws of quantum physics rather than computational complexity \cite{qiao2024transitioning}. The security of QKD, in contrast to classical encryption, is based on quantum ideas like entanglement, quantum superposition, and the Heisenberg uncertainty principle. This guarantees that any attempt to eavesdrop will inevitably cause detectable disruptions to the transmitted quantum states \cite{imran2024quantum}.

Bennett and Brassard's 1984 QKD system, BB84, makes the most impactful use of polarized photons delivered in randomly selected polarization bases to create encryption keys. Any opponent trying an interception invariably betrays their presence due to quantum features such as the no-cloning theorem, which forbids accurate copying of unknown quantum states, as their measurements damage the original quantum information. Beyond BB84, other QKD systems as the Ekert protocol, use quantum entanglement to provide key security by Bell inequality violations, therefore highlighting even more the important function of quantum mechanics in safe communications \cite{rusca2024quantum}.

QKD systems provide excellent defenses against hostile data injection, interception, or illegal access, therefore addressing numerous classical vulnerabilities in conventional FL environments in the scope of QFL. Typically in federated systems, model updates have to be safely sent from edge devices to a central server. Given that the quantum-generated keys are theoretically immune to both classical and quantum attacks \cite{ravikumar2023quantum}, QKD guarantees that these communications stay private. Their combined optimization not only allows adaptive encryption techniques depending on real-time threat levels and network conditions, but also helps QKD be included in QFL. QFL can, for example, dynamically change model update rates or communication intervals depending on the key refresh rate or channel noise in QKD. This co-design enhances not only security but also system performance, therefore strengthening the whole framework and increasing its efficiency.

Furthermore, providing resilience against possible future threats presented by quantum computing itself, quantum-secured FL protects data integrity even against quantum-enabled enemies. Integrating QKD into QFL thus marks a major step toward really safe, quantum-resistant machine learning systems. It uses the physical rules of quantum mechanics to offer a strong security layer, therefore greatly increasing trust and privacy in distributed learning systems and opening the path for future quantum-proof secure communications in federated systems.

\begin{figure}[htbp]
\centerline{\includegraphics[width=0.99\linewidth]{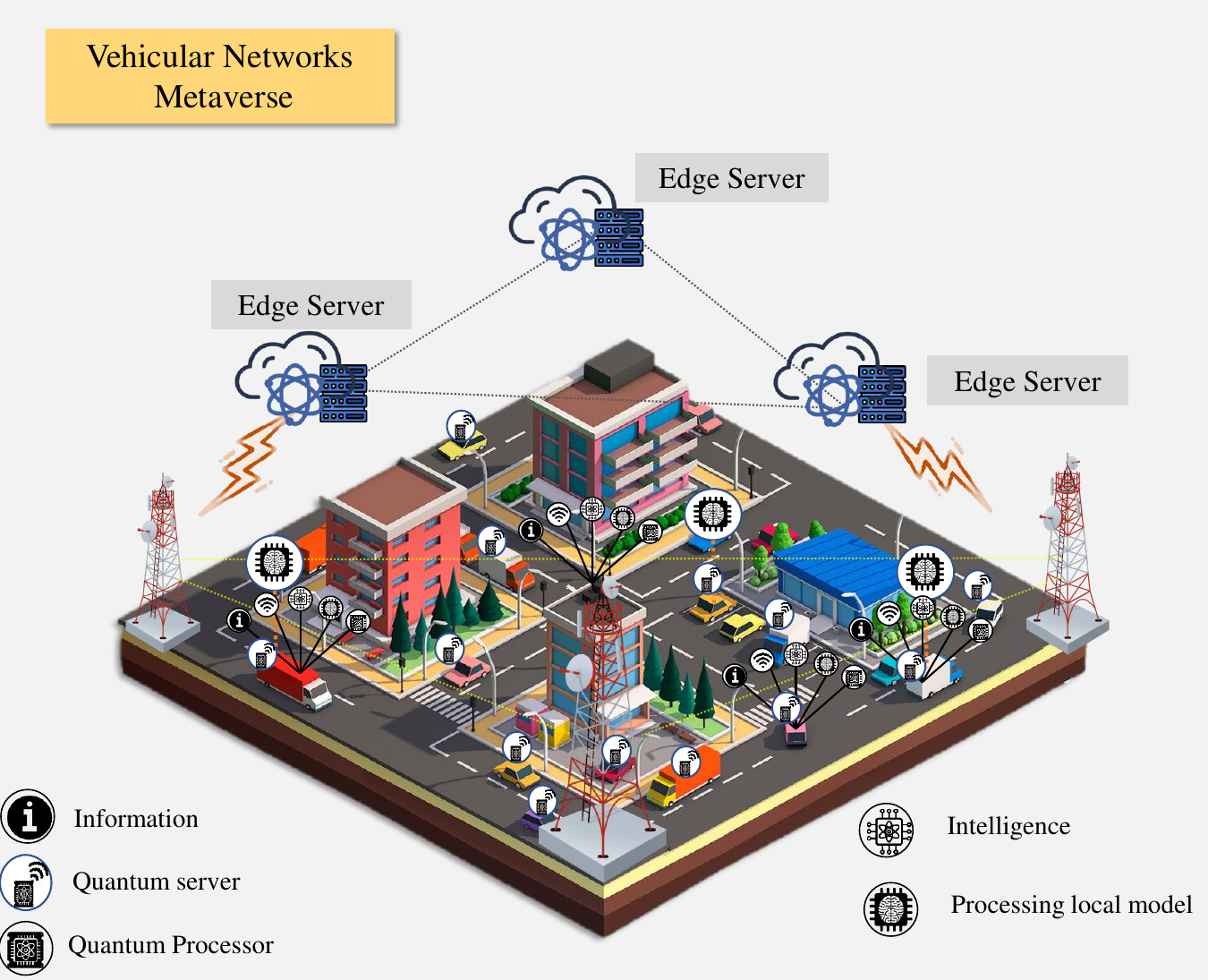}}
\caption{Detailed Architecture of QFL Framework for Vehicular Metaverse Environments: Integration of Edge Servers, Quantum Processors, and Intelligent Agents for Secure and Scalable Information Processing and Local Model Training in Urban Vehicular Networks}
\label{qfl_vehicular}
\end{figure}

\subsubsection {Quantum Homomorphic Encryption}

QFL presents safe and privacy-preserving computations over remote nodes by including quantum computing methods in FL. Among the fundamental security systems, Quantum Homomorphic Encryption (QHE) stands out as a useful tool offering quantum analogs to classical homomorphic encryption. Classical homomorphic encryption preserves privacy and confidentiality during processing by letting calculations be performed on encrypted material without ever decrypting. QHE allows quantum computations straight on encrypted quantum states without compromising security, hence extending this ability into quantum computer domains \cite{zhang2021multi}.

Fundamentally using quantum physics ideas, including quantum superposition, entanglement, and quantum uncertainty, QHE improves data confidentiality. QHE guards quantum data processing against quantum-powered attackers, unlike classical homomorphic encryption that depends on computational assumptions vulnerable to quantum computational attacks (e.g., Shor's method). First theoretically proposed by Liang in 2013, QHE systems preserve the coherence and encryption of quantum encrypted states by directly applying quantum gates onto them \cite{hu2023privacy}.

By means of QHE in QFL systems, remote nodes can safely provide encrypted quantum data or quantum-encoded classical data to a central quantum-activated server. After that, the server cooperatively trains advanced quantum-enhanced machine learning models by quantum calculations on these encrypted states without decrypting individual data entries. This strategy greatly reduces privacy concerns and stops illegal data flow. Recent studies show that quantum homomorphic encryption is especially fit for federated systems, including sensitive data such as banking or healthcare, where privacy protection is crucial \cite{kannan2024quantum}.
Nevertheless, given current quantum computing constraints, including maintaining coherence, controlling quantum noise, and obtaining adequate quantum gate authenticity, effective deployment of QHE in QFL is difficult. Also, these challenges, through continuous study, are gradually overcoming theoretical and practical constraints, improving quantum processing performance and the practical relevance of QHE. Thus, a major development in quantum-secured FL is quantum homomorphic encryption, which enables safe cooperative quantum computations and offers a strong defense against upcoming quantum adversaries.

\subsubsection {Quantum Differential Privacy}
QFL employs quantum computing principles to enhance the privacy and security of FL models, and one prominent approach is Quantum Differential Privacy (QDP). Differential privacy, traditionally a classical concept, is designed to protect individual data contributions by ensuring that the removal or addition of a single data point does not significantly alter the aggregated statistical outputs, thus providing robust privacy guarantees. Quantum differential privacy expands this classical notion into the quantum computing paradigm, integrating quantum mechanics and quantum information theory to safeguard sensitive quantum or classical data processed in quantum-enabled FL systems \cite{hirche2023quantum}.

Additionally, QDP involves carefully perturbing quantum states or quantum measurements to limit the disclosure risk about individual quantum states or classical data encoded quantumly. This perturbation typically leverages quantum uncertainty, quantum noise, and probabilistic quantum measurement outcomes to ensure strong theoretical privacy guarantees. QDP by adapting classical definitions to the quantum domain, introducing a privacy mechanism that employs quantum noise through carefully designed quantum channels to achieve privacy protection \cite{zhao2024bridging}. By integrating such quantum mechanisms within QFL, participants collaboratively train machine learning models using quantum computations while ensuring individual contributions remain indistinguishable and secure from adversaries.

Implementing quantum differential privacy (QDP) in QFL frameworks involves encoding sensitive classical or quantum data into quantum states and applying privacy-preserving quantum noise operations, such as quantum gates or quantum channels, before aggregation or measurement. These operations introduce uncertainty, measurement disturbance, and inherent noise, which make it significantly harder for adversaries to infer individual data points from learning outcomes. Recent studies demonstrate that QDP can provide stronger privacy guarantees compared to classical differential privacy, particularly in quantum settings. For instance, as shown in Table I of \cite{kannan2024quantum}, the privacy score of the QFL system is as high as 0.97, which is significantly higher than the traditional benchmarks of 0.80 to 0.90. The improved privacy score, in conjunction with a high resilience of 0.95 and communication security of 0.93, illustrates that QDP not only enhances privacy but also maintains robust overall system performance in the face of quantum threats.

\subsubsection {Blind Quantum Computing}
Within QFL, Blind Quantum Computing (BQC) is a fundamental security technique that allows devices to assign quantum computations to strong quantum servers without disclosing sensitive quantum data or computational tasks. Originally presented by Broadbent et al. in 2009, BQC ensures that the quantum server running the computations hides completely the computational input, intermediate quantum states, and final output \cite{ma2024universal}.

Blind Quantum Computing is fundamentally based on the intrinsic uncertainty and unpredictability of quantum states, mostly obtained by quantum superposition, entanglement, and cryptographic obfuscation using randomly generated quantum states. Along with classical instructions encrypted using quantum principles, the customer generates qubits in randomly chosen quantum states and safely delivers them to a quantum server. Ignorant of the precise preparation or computational directions of the quantum states, the server executes quantum gates as directed and generates measurement results \cite{amoretti2021private}. The device then decodes the results to make sure the server learns nothing significant about the kind or outcome of the computation.

Particularly against malicious quantum servers, including BQC into QFL, increases privacy inside distributed learning systems. Devices using FL safely offload quantum computational activities via BQC protocols, therefore safeguarding both quantum and classical training datasets. BQC thereby protects the collaborative training process, allowing strong quantum-enhanced machine learning without violating individual data privacy \cite{innan2024fedqnn}. With pragmatic difficulties like quantum coherence maintenance and noise management, continuous research helps Blind Quantum Computing scalability and dependability to be improved, thus stressing its vital importance in safe QFL.

\begin{table*}
    \centering
    \caption{ Comparison of solutions addressing different QFL fundamentals.}
    \label{tab:qfl_fundamentals_cos}
    \small
    \begin{adjustbox}{max width=\textwidth}
    \begin{tabular}{|m{3.2cm}|m{2.3cm}|m{9.8cm}|m{2cm}|}
    \hline
         \textbf{Taxonomy} & \textbf{Fundamentals} & \textbf{Defination and Solution Analyzed in the Literature} & \textbf{References}\\
         \hline
         
         {{\makecell{ Federation Architecture \quad}}} & Fully QFL & \begin{itemize} \item Each node maintains entirely quantum-based models, typically built using PQC. \item Approach leverages quantum parallelism and coherence, significantly improving computational efficiency.\end{itemize} & \cite{huang2022quantum, abou2024privacy, park2025entanglement, innan2024fedqnn} \\ \cline{2-4}
         & Hybrid QFL & \begin{itemize} \item Combines classical neural network layers with quantum layers. \item Improved scalability and realistic deployment by blending classical computing robustness with quantum computing expressivity. \end{itemize} & \cite{hisamori2024hybrid, abbas2021power, jaksch2023variational, ren2023qfdsa}\\ \hline
         
         {{\makecell{Networking Topology \quad}}} & Centralized QFL & \begin{itemize} \item Employs a single quantum server aggregating quantum or classical model updates from multiple distributed quantum clients. \item A critical point susceptible to single-point failures and potential bottlenecks.\end{itemize} & \cite{qiao2024transitioning, singh2021quantum, chehimi2023foundations}\\ \cline{2-4}
         & Hierarchical QFL & \begin{itemize} \item A multi-layered network structure where quantum clients communicate with quantum edge servers for local aggregation before forwarding updates to a central quantum cloud. \item This multi-tiered structure significantly improves scalability, reduces latency, and mitigates communication bottlenecks compared to centralized approaches.\end{itemize} & \cite{su2025joint, abad2020hierarchical, lim2021decentralized}\\ \cline{2-4}
         & Decentralized QFL & \begin{itemize} \item Adopts a peer-to-peer communication topology. \item Enhances fault tolerance, scalability, and privacy protection.\end{itemize} & \cite{kannan2024quantum, malina2021post, ghaderibaneh2022efficient}\\ \hline

         {{\makecell{Communication Schemes \quad}}} & Classical Communications & \begin{itemize} \item Involve exchanging classical model parameters using traditional wireless or wired channels such as 5G/6G networks or Wi-Fi. \item Ensure broad accessibility and robustness for practical QFL deployment.\end{itemize} & \cite{wang2023quantum, park2024dynamic, quy2024federated}\\ \cline{2-4}
         & Quantum Communications & \begin{itemize} \item Transfer quantum states or quantum-encoded information via specialized quantum channels. \item Inherently offers unprecedented security through quantum principles like entanglement and quantum uncertainty.\end{itemize} & \cite{yuan2020free, abou2024privacy, abou2012power, vu2020design}\\ \hline
         {{\makecell{Optimizations for QFL \quad}}} & Model Optimization for QFL & \begin{itemize} \item Efficiently training quantum or hybrid quantum-classical models, primarily through algorithms tailored to PQC. \item Mitigate barren plateau issues and quantum noise, significantly improving training effectiveness. \end{itemize} & \cite{qi2024federated, qi2023optimizing, liu2022layer, yulianti2022implementation}\\ \cline{2-4}
         & Communication Optimization & \begin{itemize} \item Focuses on reducing the overhead associated with exchanging model updates across quantum nodes, employing compression and adaptive aggregation methods. \item  Selective updates, and adaptive aggregation schedules significantly reduce communication bandwidth and latency, facilitating efficient scaling of QFL deployments.\end{itemize} & \cite{xu2022privacy, bhatia2024communication, kaewpuang2023adaptive, narottama2023federated}\\  \hline

         {{\makecell{Security Mechanisms \quad}}} & Quantum Key Distribution & \begin{itemize} \item QKD securely generates cryptographic keys between quantum nodes based on quantum mechanics principles, enabling inherently secure communication. \item Effectively addresses interception risks by leveraging quantum uncertainty. \end{itemize} & \cite{qiao2024transitioning, imran2024quantum, rusca2024quantum, ravikumar2023quantum}\\ \cline{2-4}
         & Quantum Homomorphic Encryption & \begin{itemize} \item Directly on encrypted quantum states, preserving privacy without data decryption. \item Robustly protects sensitive data during distributed quantum computations. \end{itemize} & \cite{zhang2021multi, hu2023privacy, kannan2024quantum}\\ \cline{2-4}
         & Quantum Differential Privacy & \begin{itemize} \item Provided a basic overview and the potential of QFL. \item Strongly protects against privacy breaches in federated quantum training. \end{itemize} & \cite{hirche2023quantum, zhao2024bridging, kannan2024quantum}\\ \cline{2-4}
         & Blind Quantum Computing & \begin{itemize} \item Enables clients to delegate quantum computations securely to quantum servers without revealing inputs. \item Approach robustly safeguards quantum data and computational integrity against server-side adversaries.\end{itemize} & \cite{ma2024universal, amoretti2021private, innan2024fedqnn}\\ \hline
    \end{tabular}
    \end{adjustbox}
\end{table*}

\textcolor{black}{Providing fundamentals of QFL from its diverse architectures and communication schemes to its robust security mechanisms, we now turn our attention to its practical impact. The theoretical constructs discussed in this section serve as the essential toolkit for deploying QFL in real-world scenarios. The following section will explore how these fundamental concepts are being applied across various innovative domains, demonstrating the tangible benefits and transformative potential of QFL in solving critical industry challenges.}

\section {Applications of QFL} \label{applicationsofqfl}

\color{black}
Building on the fundamental components and taxonomies established in the previous section, we now shift our focus from the theoretical blueprint to practical implementation. This section explores the transformative impact of QFL across a diverse range of real-world domains. We will examine how QFL is leveraged to solve critical challenges and unlock new capabilities in areas from vehicular and satellite networks to healthcare and the metaverse. For each domain, we discuss current state-of-the-art implementations and conclude by synthesizing the key challenges to outline specific future research directions. A summary of these applications can be found in Table~\ref{tab:security_application}.
\color{black}
\subsection { QFL for Vehicular Networks}
The vehicular metaverse needs an FL structure that is decentralized, based on quantum mechanics, and aware of heterogeneity. Quantum-based decentralized and heterogeneity-aware federated learning framework for vehicular metaverse (QV-FEDCOM) combines advanced quantum computing ideas with FL to handle the changing and unique needs of vehicle networks well. A quantum sequential training program with dynamic mode switching and a vehicle-context grouping system are two important parts that work together to make communication and data handling faster and better. Using QV-FEDCOM is a big step forward in solving the tricky problems of the vehicle metaverse, and it looks like it will make things run better and be more flexible \cite{hazarika2024quantum}.

In fact, dynamic QFL to enhance vehicular computing and public safety by managing autonomous vehicles more effectively. The DQFL framework integrates quantum computing into FL to address the challenges posed by the increasing scale and data demands of AV networks. It particularly focuses on real-life scenarios, where AVs use QNNs for tasks such as image classification, which is crucial for road safety tasks such as license plate recognition. The introduction of DQFL aims to revolutionize vehicular computing by optimizing AVs' data processing and task execution capabilities, promising significant advancements in public safety and vehicular management \cite{kim2023quantum}. Also, a cutting-edge quantum-based FL framework is designed specifically for the vehicular metaverse. QV-MetaFL combines quantum computing with FL to tackle the unique challenges of vehicular networks, such as data heterogeneity and communication efficiency. Key components include the quantum sequential training program and the quantum vehicle context grouping, which together enhance the learning process and data management. QV-MetaFL marks a significant advancement in FL, showing promising results in simulations that underline its potential to transform vehicular computing in the metaverse \cite{hazarika2024quantumc}.
Here, Equipping every vehicle and infrastructure component within the Internet of Vehicles (IoV) with quantum processors is considered impractical \cite{chougule2024exploring}. The concept of the vehicular metaverse presents a strategic alternative, focusing on managing quantum computations at the edge or within virtual environments using dedicated quantum servers. This approach optimizes the integration of quantum computing into vehicular networks, making it more feasible and cost-effective \cite{sood2024scientometric}.

In \cite{chehimi2024fundamentals}, the researcher illustrates, building on the foundation of integrating quantum computing with vehicular networks, QFL emerges as a significant advancement over Classical FL. By leveraging QNNs, QFL enhances data processing capabilities, model precision, and security, aligning well with the dynamic needs of the vehicular metaverse. In Fig. ~\ref{qfl_vehicular}, the transition to QFL represents a crucial step, not just incremental, in keeping pace with rapid technological advancements and ensuring that vehicular systems do not become obsolete. In particular, the innovative federated QNN introduces quantum teleportation to optimize resource allocation in wireless communications, showing enhanced efficiency and effective power management in NOMA-based systems \cite{shahjalal2023enabling}. Here, research highlighted in recent studies explores a QFL-based approach that safely and effectively combines local model parameters using quantum bits, applicable in both centralized and decentralized settings.

However, the authors in \cite{yamany2021oqfl} propose an optimized QFL framework aimed at improving the security of AVs by optimizing FL hyperparameters against adversarial assaults. Highly robust to adversarial threats like data poisoning, the OQFL framework dynamically adjusts learning rates and the length of local and global training epochs using a quantum-behaved particle swarm optimization technique. This method not only strengthens the protection mechanism inside the AV ecosystem but also, based on assessments using the MNIST and Fashion-MNIST datasets, greatly improves accuracy compared with related methods. All things considered, OQFL offers a safer and effective way for managing private data in distributed situations, hence redefining distributed machine learning in AVs.

Meanwhile, the authors \cite{xu2023secure} examine the integration of quantum communication technologies with FL in autonomous vehicular networks, specifically quantum AVNs. The study develops a system architecture for QAVNs using quantum key distribution algorithms and a space-air-ground integrated network model to improve data security and privacy. The main technical paper proposes a safe FL system using local differential privacy and homomorphic encryption to prevent inference attacks, eavesdropping, and Sybil attacks. This study improves FL security in QAVNs and suggests ways to improve FL efficiency and privacy in autonomous driving.

\subsection {QFL for Healthcare Networks}
The Dynamic Aggregation Quantum Federated Learning (DAQFL) technique is intended to improve intelligent diagnosis within the Internet of Medical Things. DAQFL tackles the issues of data heterogeneity in medical settings using a dynamic weighted aggregation approach that modifies the impact of device data according to real-time performance feedback. This method enhances the precision of the global model by alleviating the impact of data bias. The DAQFL technique employs QNNs and constructs VQCs with significant entanglement capabilities, thereby improving accuracy, training speed, privacy protection, and resistance to noise \cite{qu2025daqfl}.

The authors \cite{tanbhir2025quantum} introduce a quantum-inspired encryption-based FL system for dementia classification to improve healthcare privacy and security. The method uses QKD to protect model weights during federated CNN training among distant healthcare nodes. QKD protects critical patient data from gradient inversion and eavesdropping. This quantum-enhanced FL method makes AI-driven diagnostics scalable and secure for resource-constrained settings in low- and middle-income nations, boosting quantum medical research.

Besides, an innovative architecture that combines edge computing, quantum transfer learning, and FL to transform pain level evaluation through ECG data processing. This method centers on converting one-dimensional ECG data into two-dimensional Continuous Wavelet Transform pictures, which are then analyzed using a Quantum Convolutional Hybrid NN to improve feature detection and classification. This revolutionary approach represents a substantial improvement in pain assessment, ensuring more precise, secure, and efficient patient care within healthcare technology \cite{balasubramani2025novel}.

Here, to improve data privacy in healthcare, this research presents a new FL system including quantum tensor networks (QTNs). To maximize the training of models on distant datasets, the framework uses ideas of many-body quantum physics, hence lowering the demand for significant parameter use and communication bandwidth. This method not only guarantees sensitive data using differential privacy analysis but also exceeds conventional FL models in accuracy and efficiency. All things considered, the deployment of QTNs in federated environments promises notable improvements in safely and effectively managing private healthcare data \cite{bhatia2024federated}.

Another research work proposes a Hybrid Federated Learning (HFL) architecture designed to protect privacy in applications related to mental health. Clustered Federated Learning (CFL) and QFL are combined in the framework, whereby QFL uses a variational quantum classifier for advanced data categorization and CFL improves device-specific learning characteristics. By means of extensive testing, QFL demonstrated remarkably high accuracy and recall improvements, whereas CFL exhibited notable increases in precision above conventional FL techniques, therefore illustrating the efficacy of both methods \cite{gupta2024privacy}.

To transform individualized medicine, the authors in \cite{mondal2024ai} present an artificial intelligence-driven Big Data Analytics Framework that specifically blends FL, blockchain, and quantum computing. Using FL for safe, localized data handling, Blockchain for open and safe data transactions, and Quantum Computing for fast data processing speeds and strong encryption, the system improves data privacy, scalability, and computational efficiency. Demonstrating superior performance in diagnostics and patient data management, this architecture promises significant advancements in healthcare by improving diagnostic accuracy, treatment efficiency, and data security, setting a new standard for future medical technologies.

By offering much improved performance, quantum computing is a developing technology that could revolutionize data processing and optimization, especially in healthcare analytics. The possibility of quantum computing to challenge current cryptographic systems calls for the development of quantum-resistant algorithms, so enhancing safe data exchange in uses connected to health. Furthermore, the integration with big data and artificial intelligence systems might transform healthcare data administration, thereby improving patient diagnosis and prognosis prediction. These technical developments, which include artificial intelligence, big data analytics, FL, blockchain, and quantum computing, taken together, are reshining personalized medicine \cite{jeyaraman2024revolutionizing}.

On the other hand, important for estimating hepatic steatosis levels in liver transplantation, this work develops sophisticated algorithms to improve the categorization of liver biopsy images. Utilizing quantum machine learning for its superior generalization capabilities, the research introduces a hybrid quantum neural network model that significantly boosts diagnostic accuracy.  Moreover, to address privacy issues in sharing patient data between hospitals, the study employs a privacy-conscious FL approach. Also, this work presents a scalable and stable method that aids clinical pathologists in enhancing diagnostic efficiency and accuracy in evaluating liver diseases \cite{lusnig2024hybrid}.
Some recent studies show how well artificial intelligence diagnoses liver illnesses, especially via picture classification jobs where machine learning shines. These models need confirmation by extensive clinical trials, even if their outcomes match those of experts. Furthermore, a Hybrid Quantum Convolutional Neural Network (HQNN) is promising in brain tumor classification and displays faster convergence than conventional models. Nevertheless, the sparse availability of quantum computers and the difficulties in modeling quantum processes on classical systems restrict the development of QML in medicine \cite{wang2025hqnet}.

The authors \cite{qu2024quantum} present, the focus is on data privacy in quantum environments, QFL employing quantum data and processors for both devices and servers. Their results imply that quantum-inspired approaches might be included in classical FL systems to improve security, particularly in delicate fields like dementia detection. This concept might combine quantum techniques with conventional encryption to greatly increase data security in classical systems.

 Lastly, maintaining privacy and security of medical data is critical in the developing field of healthcare, especially in sensitive areas like dementia categorization.  Inspired by quantum computing, a Quantum-Inspired Privacy-Preserving FL framework improves the security and efficiency of FL systems.  Using quantum cryptography methods like quantum random number generators (QRNGs) and QKD, this system guarantees the transfer of model changes over collaborating nodes. The QFL framework guarantees that every update is securely encrypted and that data integrity is kept across the learning process by combining these quantum elements. This not only guards private and sensitive patient information from possible cyberattacks but also improves the collaborative learning process by enabling the development of more accurate and private dementia classification models over dispersed healthcare networks \cite{park2024dynamic}.


\subsection {QFL for Satellite Networks} 

\begin{figure}[htbp]
\centerline{\includegraphics[width=0.99\linewidth]{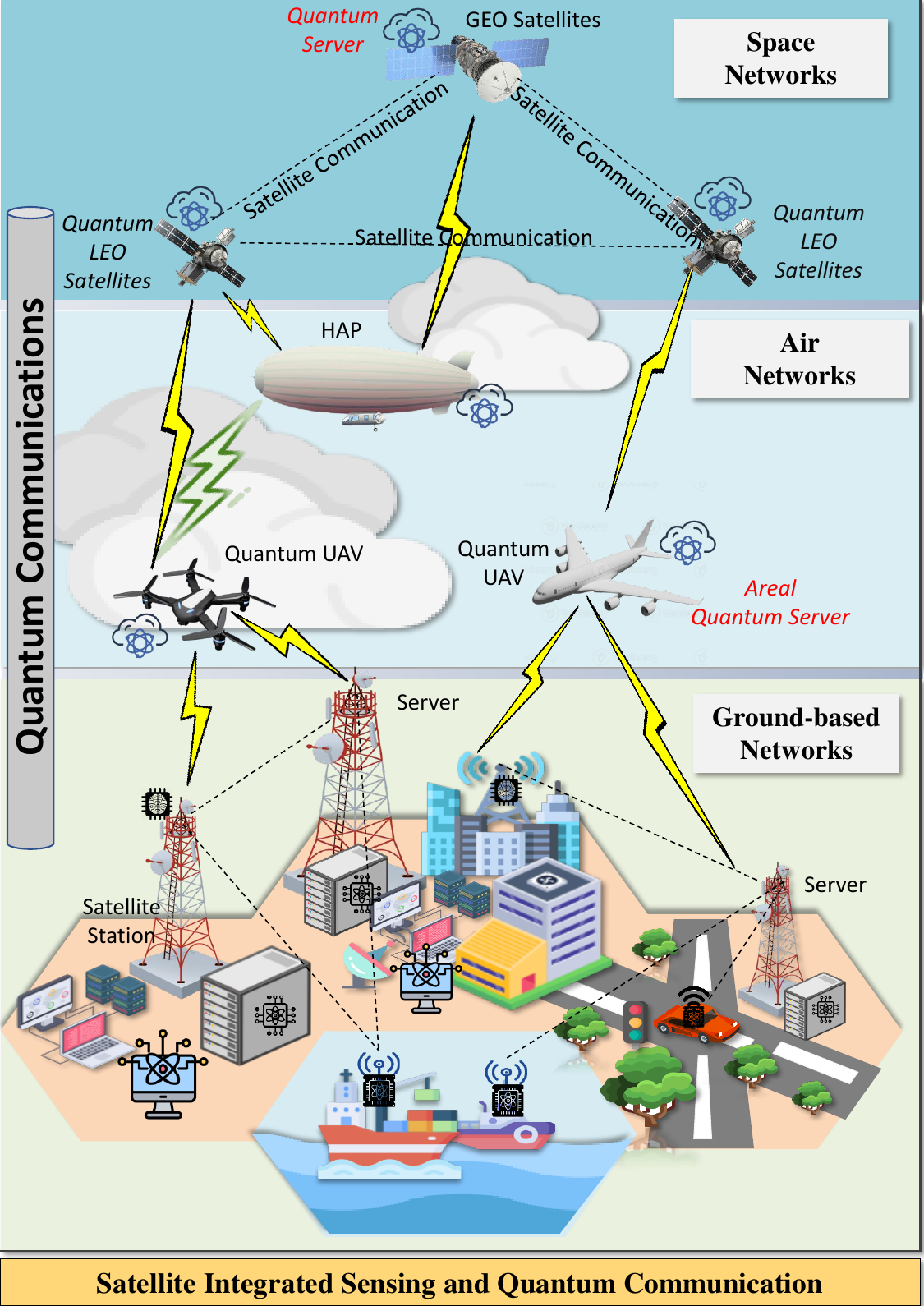}}
\caption{The illustration of satellite-ground QFL framework. It consists of three processes: (1) Ground-Based Networks, (2) Air-Based Networks, and (3) Space Networks.}
\label{qfl_satellite}
\end{figure}

\textcolor{black}{QFL was used for the first time in satellite-ground communication, utilizing the capacity of LEO satellites to cover the entire planet. With adjustable configurations (angle and pole), it suggests slimmable quantum federated learning (SQFL) and slimmable quantum neural networks (sQNN) that enhance communication and processing. The possibilities for communication are further expanded by superposition coding and subsequent decoding.} These authors \cite{park2024dynamic} present the satellite-ground SQFL framework greatly exceeds conventional FL and QFL in efficiency, establishing a new benchmark for satellite communication systems. The illustration is done in Fig.~\ref{qfl_satellite}. The authors of \cite{ren2025toward} also state that standardizing in QFL is necessary for the successful integration of quantum computing with FL across many fields. Establishing protocols for quantum device communication, guaranteeing network compatibility, and defining security mechanisms for quantum data and model exchanges constitute part of this process. Furthermore, improving system dependability by standardizing error correction for quantum systems solves fundamental problems such as quantum noise and decoherence. Setting these criteria will help to enable strong, safe, and effective quantum-enhanced FL systems by enabling the wider acceptance of QFL in areas such as telecommunications.

In this case, a QFL architecture for SAGIN, which is a key part of the new 6G networks. It handles big datasets for training machine learning models on devices on Earth using VQA and quantum relays, which makes the data safer. Local training is done with VQA, and UAVs and high-altitude platforms (HAPs) are used to send safe quantum relay-based models over long distances. A case study shows that the VQA-based FL structure and quantum relay system can be used in real life and work well. With quantum approaches and FL in SAGIN, 6G networks could have edge intelligence systems that work well and are safe \cite{wang2023quantum}.

Moreover, the authors in \cite{alchieri2021introduction} investigate the integration of quantum deep learning into the fundamental layer of Quantum Federated Virtual Reality (QFVR) inside SAGIN, where devices endowed with quantum capacity perform local training. By amplitude encoding and other approaches, the technical work is to transform local datasets into quantum states and to build a local quantum model by parameterizing the quantum circuit manipulation of these states. By methodically reducing a traditionally stated cost function with quantum computing methods, the approach improves the model and therefore increases the security and efficiency of data processing in SAGIN.

In \cite{lv2024safeguarding}, the main methods applied include the no-cloning theorem, entanglement-based communication, and the inherent randomness of quantum measurement outcomes. The no-cloning theorem ensures that exact copying of an unknown quantum state is impossible, thereby protecting the FL model during transmission by producing essentially flawed copies. Entanglement-based communication greatly enhances security; any interception effort disturbs the entanglement, which the network can quickly identify as possible eavesdropping. The stochastic character of quantum measurements adds another level of security since the outcomes of these measurements are intrinsically erratic and so complicate any attempts to decrypt the conveyed information. Lastly, quantum methods taken together provide a strong architecture for safe model transfer in FL applications within SAGIN, therefore greatly enhancing data protection against eavesdropping and several security issues.

Lastly, the authors \cite{quy2024federated} present that the future 6G networks depend on SAGINs and the integration of FL and QFL.  The goal is to improve these networks using artificial intelligence methods to provide intelligent, real-time applications in many spheres, including military operations and emergency response. Adopting FL and QFL helps to solve issues of large data, privacy, and security, thereby improving resource use, lowering communication latency, and increasing energy efficiency in SAGINs. The benefits of the quantum-activated training strategy over conventional FL techniques are shown by a case study on UAV networks. The paper also addressed research difficulties and the requirement of standardizing QFL implementation to underline its potential to greatly change future 6G networks utilizing improved processing capacity and secure communications.

\subsection {QFL for Metaverse}

To build the Metaverse, a virtual space for business and social activities, we need systems that ensure safety, accessibility, and reliability. To meet these goals, a distributed QFL architecture has been made that uses blockchain technology \cite{khalid2023quantum}. Blockchain QFL (BQFL) makes the system safer and more open, which makes it more resistant to hacking and dishonest behavior. The BQFL system distributed design helps to lower the risks that come with having a central server. A lot of research, both theoretical and actual, has shown that BQFL is useful in the Metaverse, which is a hybrid metaverse made possible by a metaverse observer and a world model \cite{gurung2023decentralized}.

In addition, this research presents QV-FEDCOM, a novel quantum-based, distributed, and heterogeneity-aware FL system especially intended to address the special difficulties of this dynamic environment in the fast-changing domain of the vehicular Metaverse. By combining quantum computing ideas with FL, QV-FEDCOM offers a flexible and affordable solution improving the interface and functioning of the vehicle Metaverse. Furthermore, the application of quantum-inspired main component analysis (Q-PCA) maximizes memory economy, hence strengthening the Metaverse architecture. These components reinvent learning dynamics and enable the sophisticated needs of the vehicle Metaverse, therefore guiding QV-FEDCOM as a pillar for next advancements in this virtual world \cite{hazarika2024quantum}.

Moreover, the authors presents QV-MetaFL, a quantum-based FL system especially intended for the vehicle Metaverse.  Featuring major technologies including the quantum sequential-training-program (Q-STP) and quantum vehicle-context-grouping (Q-VCG), this novel system blends quantum computing with FL to effectively manage vehicle data and communications. These solutions affordably maximize learning processes and solve data heterogeneity problems. All things considered, QV-MetaFL greatly improves the dynamics of learning in the vehicle Metaverse, therefore demonstrating a major advancement in the evolution of virtual cars settings \cite{hazarika2024quantumc}.

This research integrates IoT devices and UAVs to enhance real-world metaverse synchronization. To keep data consumption minimal when base station resources are limited, offload UAV tasks to Web3 cloud servers. The authors also presents a quantum collective reinforcement learning technique that enables UAVs to quickly adjust to new environments, hence enhancing their performance in dynamic conditions. Through effective data flow and processing management, this method improves Metaverse integration between virtual and real-world experiences \cite{wang2023efficient}. Also, for realistic metaverse systems, this research introduces a new coordination method aiming at improving synchronization between the physical world and the virtual meta-space. Starting with QMARL to optimize data collecting in physical space for low-delay temporal synchronisation, uses avatar popularity for creating virtual content under delay constraints, and employs caching strategies based on avatar popularity and age-of-information to build a realistic meta-space effectively. From lowering computational needs to optimizing reality quality in virtual environments, each stage is intended to efficiently manage synchronizing issues, thereby demonstrating a complete method to create immersive and responsive metaverse systems \cite{park2024joint}.

Finally, the authors improve AI models for self-driving and connected cars by using quantum group learning and a many-to-many matching game-based method in a made-up metaverse. The metaverse's time clock mechanism improves sampling, which fixes the problem of not getting enough data when bad weather happens less often. This method emphasizes the variety of features across different types of vehicle data. This makes it possible to use multiple data forms from different vehicle brands more effectively, which improves group learning. This method uses the Gale-Shapley algorithm to play a game of many-to-many matches to get the most vehicles spread out. Spectrum resource allocation is also looked at as a discrete Markov decision process with a quantum-inspired reinforcement learning method to find the best way to make the most money for the system \cite{ren2022quantum}.


\begin{table*}
    \centering
    \caption{Taxonomy of contemporary QFL applications.}
    \renewcommand{\arraystretch}{1.5}
    \label{tab:security_application}
    \begin{adjustbox}{max width=\textwidth}
    \begin{tabular}{|>{\centering\arraybackslash}m{1cm}|>{\centering\arraybackslash}m{1cm}|>{\centering\arraybackslash}m{2cm}|>{\centering\arraybackslash}m{2cm}|>{\centering\arraybackslash}m{2cm}|>{\centering\arraybackslash}m{1cm}|>{\centering\arraybackslash}m{4cm}|>{\centering\arraybackslash}m{4cm}|}
        \hline
        \textbf{Issue} & \textbf{Ref.} & \textbf{Topic} & \textbf{Federation Architecture} & \textbf{Communication Schemes} & \textbf{Security} & \textbf{Key contributions} & \textbf{Limitations} \\ \hline
        
         \multirow{3}{*}{\rotatebox[origin=c]{90}{\makecell{QFL for  \\ Vehicular Networks \quad}}} & \cite{hazarika2024quantum} & Framework tailored for the vehicular metaverse. & Decentralized & Quantum channel & Yes & Dynamic Mode Switching, Vehicle-Context Grouping. & Focuses solely on Complexity and scalability. \\ \cline{2-8}
         & \cite{kim2023quantum} & Dynamic QFL for AVs & Integrated QFL System & Quantum and Classical Channels & Yes & Enhanced AV Computing, Real-Life Application & Implementation Complexity \\ \cline{2-8}
         & \cite{chehimi2024fundamentals} & QFL Advancement & Hybrid Centralized-Decentralized & Quantum Channel & Yes & Improved Quantum Teleportation, Enhanced Efficiency & Complexity in Integration \\ \cline{2-8}
         & \cite{yamany2021oqfl} & Optimized Quantum FL for AV Security & Distributed QFL & Quantum Channel & Yes & Dynamic Hyperparameter Optimization & Dependency on Quantum Optimization \\ \hline

          \multirow{3}{*}{\rotatebox[origin=c]{90}{\makecell{QFL for  \\ Healthcare Networks \quad}}} & \cite{qu2025daqfl} & Quantum Medical Diagnosis & Dynamic Aggregation FL & Quantum Channel & Yes & Weighted Aggregation, Noise Resistance & Quantum Resource Availability \\ \cline{2-8}
         & \cite{tanbhir2025quantum} & Quantum Healthcare Security & Quantum-Enhanced FL & Quantum Channel & Yes & QKD Integration, Dementia Classification & Quantum Implementation Complexity \\ \cline{2-8}
         & \cite{balasubramani2025novel} & Quantum Pain Assessment & Edge-based FL & Quantum Channel & Yes & ECG Image Analysis & Quantum Computational Demand \\ \hline

         \multirow{3}{*}{\rotatebox[origin=c]{90}{\makecell{QFL for  \\ Satellite Networks \quad}}} & \cite{park2024dynamic} & Satellite Quantum FL & Satellite-Ground FL & Quantum Channel & Yes & Slimmable QNN, Superposition Coding & Deployment Complexity \\ \cline{2-8}
         & \cite{yun2022slimmable} & QFL Standardization & Cross-Sector FL & Quantum Channel & Yes & Protocol Standardization, Error Correction & Quantum Noise Challenges \\ \cline{2-8}
         & \cite{wang2023quantum} & Quantum SAGIN & VQA-based FL & Quantum Channel & Yes & Quantum Relay, VQA Integration & Quantum Infrastructure Dependency \\ \cline{2-8}
         & \cite{alchieri2021introduction} & Quantum Virtual Reality & Quantum-based FL & Quantum Channel & Yes & Amplitude Encoding, Quantum Circuits & Quantum Resource Demand \\ \cline{2-8}
         & \cite{quy2024federated} & 6G Quantum SAGIN & Quantum-Activated FL & Quantum Channel & Yes & Real-Time Applications, Resource Efficiency & Standardization Challenges \\ \hline

         \multirow{3}{*}{\rotatebox[origin=c]{90}{\makecell{QFL for  \\ Metaverse \quad}}} & \cite{gurung2023decentralized} & Blockchain Quantum Metaverse & Distributed Blockchain FL & Quantum Channel & Yes & Blockchain Integration, Transparency & System Complexity \\ \cline{2-8}
         & \cite{hazarika2024quantum} & Vehicular Quantum Metaverse & Distributed Quantum FL & Quantum Channel & Yes & Quantum-inspired PCA & Quantum Computational Overhead \\ \cline{2-8}
         & \cite{hazarika2024quantumc} & Vehicular Metaverse FL & Quantum-based FL & Quantum Channel & Yes & Q-STP, Q-VCG & Data Heterogeneity Complexity \\ \cline{2-8}
         & \cite{park2024joint} & Metaverse Synchronization & Quantum-based FL & Quantum Channel & Yes & QMARL, Avatar-based Caching & Computational Demands \\ \cline{2-8}
         & \cite{ren2022quantum} & Quantum Autonomous Vehicles & Quantum Collective FL & Quantum Channel & Yes & Matching Game, Quantum RL & Simulation Complexity \\ \hline

         \multirow{3}{*}{\rotatebox[origin=c]{90}{\makecell{QFL for Network Security}}} & \cite{abou2024privacy} & Quantum FL IDS & Distributed Quantum FL & Quantum Channel & Yes & Efficient Intrusion Detection & Quantum Resource Dependency \\ \cline{2-8}
         & \cite{yamany2021oqfl} & Quantum Cybersecurity AVs & Distributed Quantum FL & Quantum Channel & Yes & Hyperparameter Optimization, Cyber Defense & Quantum Complexity \\ \cline{2-8}
         & \cite{subramanian2024hybrid} & Cyber-Attack Detection & Hierarchical FL & Classical Channel & Yes & Dynamic Node Grouping & Aggregation Complexity \\ \cline{2-8}
         & \cite{ren2023qfdsa} & Quantum-secured DSA & Decentralized FL & Quantum Channel & Yes & QKD Integration & Quantum Resource Constraints \\ \cline{2-8}
          & \cite{maouaki2025qfal} & Quantum Adversarial Learning & Quantum Federated & Quantum Channel & Yes & Adversarial Robustness & Robustness-Accuracy Tradeoff \\ \cline{2-8}
         & \cite{ren2024variational} & Quantum Smart Grids & Hybrid Quantum FL & hybrid channel & Yes & MDI-QKD Optimization & Implementation Complexity \\ \hline    
    \end{tabular}
    \end{adjustbox}
\end{table*}

\subsection {QFL for Network Security}
Quantum Federated Learning IDS (QFL-IDs), a new framework intended to improve IDS in consumer networks by mixing Quantum Computing with FL. By addressing the scalability and privacy concerns related to conventional AI-based IDS, this integration renders QFL-IDS a strong, reliable, and privacy-preserving solution. Using the distributed character of FL, QFL-IDS lets several consumer devices cooperatively train a global model without violating individual data privacy. Furthermore, the framework makes great use of quantum computing processing capability to greatly increase the model training and inference process efficiency. The authors show that QFL-IDS provides better detection accuracy and computing efficiency than current IDS methods, therefore showcasing its potential to properly and swiftly solve new security challenges \cite{abou2024privacy}.
Besides, a unique OQFL framework to handle the security issues inside intelligent transportation systems, especially AVs. This framework is especially intended to improve the security of AV ecosystems, which depend mostly on machine learning algorithms for operations and are prone to adversarial assaults, such as data poisoning. The OQFL system dynamically changes important hyperparameters, like the learning rate and epoch lengths, using a quantum-behaved particle swarm optimization method, therefore strengthening the FL models against such attacks. Moreover, the structure is included in a cyber defense plan to offer a strong defense against enemy challenges. Benchmark datasets helped to confirm the OQFL framework efficiency in improving security by proving its higher resilience than traditional FL methods \cite{yamany2021oqfl}.
In such contexts, a new cyber-attack detection model is proposed to improve network security by combining spatio-temporal attention networks with a quantum-inspired federated averaging optimization. The model uses a hierarchical model aggregation that dynamically groups nodes depending on network conditions and data similarity, therefore improving the adaptability of cyber-attack detection. It also includes privacy preservation methods and multi-stage model refining, thereby greatly enhancing security and performance. The authors show that the method solves the shortcomings of conventional centralized approaches, such as data privacy and communication overheads, in identifying several kinds of anomalies, so advancing digital network security \cite{subramanian2024hybrid}.

Another work presents a novel method for improving data-driven security assessments in smart cyber-physical grids: the quantum-secured federated dynamic security assessment (QFDSA) system. The system aggregates insights from many local data sources to forecast system stability by using FL and QKD, thereby decentralizing the learning process. By means of a secret-key pool and measurement-device-independent QKD, QFDSA also manages fast changes in system needs and secures data transfers. For real-time dynamic security evaluations in smart grids, QFDSA offers a scalable, safe, and effective alternative over conventional centralized approaches overall \cite{uddin2024false, ren2023qfdsa}.
By including adversarial training into QFL, this work pioneers Quantum Federated Adversarial Learning (QFAL), hence improving security against adversarial assaults. The method deliberately balances robustness and accuracy depending on device count, adversarial training degree, and attack strength by involving devices in creating local adversarial examples and aggregating these using federated averaging. Emphasizing the importance of well-calibrated adversarial training procedures in quantum federated systems, the research uncovers a fundamental trade-off between resilience and precision \cite{maouaki2025qfal}.

However, inspired to improve stability in smart grids, including renewable energy sources, this work presents the Quantum Federated Learning-based Dynamic Security Assessment (QFLDSA) technique. Crucially for grid stability evaluations, QFLDSA effectively processes high-dimensional data and complex differential-algebraic equations by combining hybrid quantum-classical machine learning with FL. With a high performance and a maximum transmission of model parameters down to up to 1000 times, the technique much exceeds conventional models. All things considered, QFLDSA provides a strong and dependable method for handling the dynamic security issues of contemporary smart grids, so opening the path for creative grid management solutions \cite{ren2024enhancing}.
Lastly, this research \cite{ren2024variational} introduces QQFL, a new hybrid QFL technique meant to maximize dynamic security assessment in smart grids. QQFL reduces the central weaknesses and inefficiencies of conventional ML-based DSA methods by merging measurement-device-independent Quantum Key Distribution with Variational Quantum Circuits, hence improving both accuracy and security. Important for the fast-evolving smart grid contexts, QQFL can perform quick online learning and dynamic deployment by using a creative use of a DNN-based MDI-QKD optimizer and Quantum Processing Units. By greatly improving security, dependability, and efficiency of smart grids, QQFL shows its promise as a strong answer for contemporary smart grid issues.

\color{black}
\subsection{Application-Specific Challenges and Research Directions}
While QFL holds immense promise across various domains, its practical realization requires overcoming domain-specific challenges. This subsection outlines key research directions tailored to the applications discussed previously, providing a roadmap for future investigation.

\begin{itemize}[leftmargin=*]
    \item \textbf{For Vehicular Networks:} The primary challenge is the highly dynamic and intermittent connectivity between vehicles \cite{du2025toward}. Future research should focus on developing \textit{asynchronous QFL protocols} that are robust to frequent client dropouts. Another key direction is designing lightweight quantum circuits and model compression techniques suitable for the resource-constrained computational hardware in vehicles \cite{liu2025distributed}. A critical research question is: \textit{How can we design quantum error correction codes that are efficient enough to maintain qubit coherence amidst the constant motion and environmental noise of a moving vehicle?}

    \item \textbf{For Healthcare Systems:} Data heterogeneity (non-IID data) is a major hurdle, as medical data varies significantly across hospitals and patient populations. A key research direction is the development of \textit{personalized or clustered QFL algorithms} that can create specialized models for different data distributions while still benefiting from collective learning \cite{qu2025daqfl}. Furthermore, ensuring the interpretability of complex quantum models is crucial for clinical adoption \cite{gurung2025performance}. This leads to the research question: \textit{Can we develop hybrid quantum-classical models that offer both the predictive power of quantum mechanics and the explainability of classical AI for trustworthy medical diagnostics?}

    \item \textbf{For Satellite Networks:} The extreme distances and harsh space environment pose significant challenges for quantum communication \cite{prakash2025quantum}. The primary research direction is to design \textit{fault-tolerant entanglement distribution protocols} for long-haul, free-space optical links. Another focus should be on creating space-resilient quantum hardware \cite{huang2025edge}. An open question is: \textit{What is the optimal trade-off between on-satellite quantum processing and transmitting quantum states to ground stations, considering the severe bandwidth and hardware limitations?}

    \item \textbf{For the Metaverse:} The Metaverse demands ultra-low latency for real-time immersive experiences \cite{gurung2025quantum}. A critical research direction is the integration of QFL with \textit{quantum-enhanced edge computing} to perform rapid model inference and updates at the network edge, close to the user. Additionally, developing QFL models for generating complex 3D assets and dynamic virtual environments is a promising avenue. A key research question is: \textit{How can QFL be used to create truly personalized and adaptive virtual experiences in real-time without compromising the privacy of user behavior data?}

    \item \textbf{For Network Security:} The main challenge is the speed at which QFL models must detect and adapt to new, zero-day threats. Future research should focus on \textit{quantum continual learning within a federated framework}, allowing security models to evolve without catastrophic forgetting \cite{qu2024quantum}. Another important direction is to explore the use of QFL for detecting adversarial attacks against the quantum machine learning models themselves \cite{subramanian2024hybrid}. This raises the question: \textit{Can we design a QFL-based intrusion detection system that is itself resilient to poisoning attacks targeting the quantum parameters of the distributed models?}
\end{itemize}
\color{black}

\section{Lesson Learned} \label{lesson}
In this section, we summarize the key lesson learned from this survey, which provide an overall picture of the current research on QFL fundamentals and applications.

\subsection{Fundamentals of QFL and Taxonomy}
The main lessons acquired from the discussions on QFL fundamentals are highlighted in the following:
 \begin{itemize}
     \item Fully QFL is a wholly quantum architecture in which every federated client preserves entirely quantum-based models, like VQE employing PQC. Under this architecture, quantum calculations are carried out straight on quantum data without first producing classical representations. Independent quantum computations by local quantum clients submit quantum or classically measured quantum parameters to a central quantum server. After that, this central aggregator distributes updated parameters and performs quantum averaging, preserving coherence and using quantum advantages for maximum processing efficiency\cite{huang2022quantum, abou2024privacy}. On the other hand, hybrid QFL combines quantum layers with classical NNs, whereby the first layers use classical NNs to preprocess and extract pertinent features, and then send input to quantum layers for further quantum transformations. By using classical computational resources together with quantum advantages, this hybridization provides pragmatic benefits that improve scalability and manageability in present NISQ systems. Nevertheless, hybrid QFL adds complexity since careful interface between classical and quantum layers is required, and improved algorithms for effective parameter mapping and optimal training procedures are thereby necessary \cite{ park2025entanglement, innan2024fedqnn}.
     \item Simplified hub-and-spoke network topology is used in centralised QFL. Here, several quantum clients interact directly with a single, strong central quantum server. Following safe quantum processing, this centralized node accumulates quantum model updates from clients and delivers the global model parameters. This architecture is fragile despite reduced communication management because of possible communication bottlenecks and single-point failure hazards \cite{hisamori2024hybrid, abbas2021power}. By comparison, Hierarchical QFL presents a multi-tiered design motivated by traditional hierarchical FL. Clusters of quantum clients interact with intermediate quantum edge computers \cite{jaksch2023variational, ren2023qfdsa}. These edge nodes aggregate initially and then transfer aggregated quantum updates to a top-tier quantum cloud server for global aggregation. Hierarchical organization improves fault tolerance, lowers latency, and helps to ease communication congestion by means of scalability. Finally, Decentralized QFL uses a peer-to-peer network topology, therefore completely removing the need for central servers. Direct exchanges of quantum model updates between quantum clients with surrounding peers offering better resilience against network failures, data privacy, and scalability for large-scale quantum networks, decentralized quantum consensus algorithms, or secure gossip protocols guarantee convergence towards a global model. But this distributed topology creates synchronizing problems that call for advanced quantum synchronization and safe quantum communication systems\cite{kannan2024quantum, malina2021post, ghaderibaneh2022efficient}.
     \item The type of the model updates shared among quantum nodes determines the communication inside QFL systems. Standard classical communication channels fit when updates incorporate classical parameters, usually numerical values received from quantum observations. Reliable and fast transmission made possible by conventional wireless technologies, including Wi-Fi or cellular networks (5G/6G), helps to enable extensive integration into current infrastructure. But when updates consist of quantum states themselves, conventional transmission falls short \cite{qiao2024transitioning, singh2021quantum}. Then quantum communications must be used, making advantage of quantum-specific transmission technologies such as fiber-optic cables or FSO networks. Using QKD, entanglement-based communication, or quantum teleportation, these quantum channels provide safe, coherent, and interference-resistant transfer of quantum states. Because of basic quantum physics ideas like the no-cloning theorem and quantum uncertainty, quantum communication systems naturally provide better security against conventional interception efforts. Nonetheless, actual implementation presents major difficulties like quantum decoherence over distances, restricted channel capacity, and complicated hardware needs, which call for continuous development in quantum communication technologies\cite{chehimi2023foundations}.
     \item In QFL, optimization methods center on improving model training efficiency, convergence rates, and communication efficacy. Mostly using VQAs, model optimization techniques consist of parameter optimization within quantum or hybrid models. The quantum-specific difficulty of barren plateaus (vanishing gradients) requires careful design of optimization strategies. Techniques including adaptive parameter initialization, noise-aware training methods, and hybrid quantum-classical optimization, leveraging classical algorithms like gradient descent or evolutionary algorithms, greatly improve model training results. On the other hand, communication optimization tackles the model update transmissions' overhead reduction. Communication efficiency is much improved by methods including selective update transmission, adaptive federated averaging schedules, and quantum model compression. Furthermore, quantum-secured aggregation techniques maximize privacy and lower pointless data interchange, therefore guaranteeing less resource consumption. Dynamic communication frequency based on client performance and network conditions helps to further enhance general resource use and scalability, hence making QFL more sensible for actual implementation.
     \item Strong quantum-enhanced security methods are demanded by QFL. Utilising quantum uncertainty and entanglement, QKD securely exchanges cryptographic keys using quantum physics. Using protocols such as BB84, eavesdropping is guaranteed, therefore shielding federated nodes against interception attacks \cite{zhang2021multi, hu2023privacy}. Direct quantum computations on encrypted quantum states made possible by QHE help to preserve data privacy during model development. QHE guarantees that quantum nodes cooperatively compute without disclosing sensitive information, hence reducing data leakage hazards inherent in distributed learning systems. Analogously, QDP uses quantum noise insertion to hide the quantum states of individual data points, hence extending conventional differential privacy notions to quantum domains. This method firmly protects participant privacy, therefore stopping adversaries from deducing private or operational data \cite{hirche2023quantum, zhao2024bridging, kannan2024quantum}. At last, BQC lets customers safely assign quantum calculations to quantum servers without disclosing quantum states or computation specifics. BQC guards computational integrity and quantum data by sending encrypted quantum inputs alongside classical instructions. These quantum security mechanisms taken together greatly improve the resilience, privacy protection, and trustworthiness of QFL systems against both quantum and classical adversaries \cite{ma2024universal, innan2024fedqnn}.
 \end{itemize}
 
 \color{black}
Synthesizing these fundamentals reveals several critical \textbf{trade-offs} and \textbf{unresolved challenges} that define the current QFL landscape. A primary \textbf{practical consideration} is the architectural choice between the near-term feasibility of hybrid QFL systems and the long-term potential of fully quantum architectures. While hybrid models are more compatible with today's NISQ-era hardware, they introduce significant overhead in managing the classical-quantum interface \cite{hisamori2024hybrid}. This leads to a key unresolved challenge: \textbf{the compatibility between classical FL protocols and quantum nodes}. For example, classical aggregation algorithms are not designed to handle quantum state vectors, necessitating new protocols that can efficiently merge quantum information without resorting to costly measurements that destroy quantum advantages. Furthermore, an inherent \textbf{trade-off} exists between the operational simplicity of centralized QFL \cite{quy2024federated} and the robustness of decentralized topologies \cite{gurung2023decentralized}. The latter introduces profound \textbf{practical challenges} in synchronizing distributed quantum states and managing entanglement across a peer-to-peer network with \textbf{emerging}, and often unreliable, quantum \textbf{communication constraints}.
\color{black}

\subsection{Applications of QFL}
The main lessons acquired from the discussions of QFL applications are highlighted in the following:
 \begin{itemize}
     \item For vehicle networks, QFL offers transforming possibilities that will allow Intelligent Transportation Systems (ITS) safe, effective, and real-time decision-making. To maximize traffic flow, route planning, accident avoidance, and autonomous driving decisions, vehicles nowadays depend more and more on distributed machine learning. By using the increased processing capability and quantum encryption features of quantum computation, QFL solves intrinsic constraints of classical FL, like latency and data security problems \cite{hazarika2024quantum}. Fast analysis of large data streams produced from sensors and vehicle-to-anything (V2X) communications by quantum-enhanced models greatly improves prediction and autonomous reaction accuracy. Furthermore, using QKD in vehicle networks guarantees safe model update transfers, therefore safeguarding important vehicle location and user behavior data against cyberattacks or illegal access. Furthermore, effective coordination of vehicle clustering by quantum-enhanced optimization algorithms improves cooperative driving and lowers communication overhead. Although present hardware limitations in quantum devices, the ongoing development toward strong quantum communications and computation systems shows the viability and significant future influence of including QFL into vehicle networks. Adoption of QFL inside vehicle environments can eventually greatly improve traffic efficiency, general network dependability, and safety \cite{chehimi2024fundamentals}.
     \item QFL presents a novel approach to safely manage private patient data in healthcare systems, hence improving computational accuracy in disease prediction, diagnosis, and tailored treatment. Large-scale, extremely private datasets created by healthcare apps are spread throughout medical facilities and hospitals \cite{qu2025daqfl}. Advanced quantum encryption available from QFL guarantees strong data protection during federated training by means of QHE and QDP. Beyond traditional computing constraints, quantum-enhanced ML models allow exact analysis of massive biomedical datasets, fast extraction of important insights linked to drug discovery, medical imaging, or patient risk assessments. Moreover, using quantum communication guarantees safe and consistent distribution of model updates, which enables hospitals to cooperatively train worldwide models without directly exchanging patient data, while carefully following privacy rules \cite{tanbhir2025quantum}. Furthermore, accelerating computationally demanding activities like molecular modeling or genetic sequence analysis is a quantum computing ability. Quantum hardware noise, limited coherence durations, and the requirement of integrating quantum-classical hybrid designs pose difficulties \cite{balasubramani2025novel}. Future developments in quantum infrastructure and algorithmic refinement could completely materialize QFL's transforming power, therefore giving healthcare practitioners fast, accurate diagnostic tools and transforming patient-centric, tailored therapy.
     \item By transcending special limitations including latency, data privacy, and security risks, QFL can significantly help satellite networks to support real-time decision-making, surveillance, and communication management. Satellites generate vast observational data, needing fast, dispersed processing \cite{park2024dynamic}. By allowing satellites to cooperatively train complex ML models, QFL quantum-enhanced algorithms help to greatly reduce the raw data transmission and therefore preserve limited bandwidth. By avoiding interception and thereby reducing risks from cyber threats inherent in satellite communications, using quantum communication links, such as FSO quantum channels, improves data security. Furthermore, quantum optimization methods enhance satellite constellation management by exactly optimizing orbital positions and lowering collision hazards \cite{wang2023quantum}. Especially in military and critical infrastructure applications, the inherent resilience of QFL against classical computational hazards guarantees the security of crucial satellite operations. Still, there are significant practical challenges for QFL deployment in satellite networks, including quantum state decoherence over space-bound transmission channels, technological limits in space-compatible quantum hardware, and consistent quantum entanglement distribution. Notwithstanding these difficulties, fast developments in quantum satellite technology and quantum communication show a bright future for extensive QFL implementation, so greatly improving satellite network security, data integrity, and efficiency \cite{quy2024federated}.
     \item Promising prospects for the growing Metaverse ecosystem, QFL addresses important issues including data security, effective distributed processing, and immersive real-time experiences, therefore tackling essential problems. Requiring strong, scalable, and safe computational resources, the Metaverse mostly depends on large-scale distributed data analysis and real-time rendering \cite{gurung2023decentralized}. By combining the parallel processing capacity of quantum computing with quantum-secure techniques, QFL greatly increases the accuracy and speed of predictive analytics and immersive simulations. Real-time, privacy-preserving interactions among geographically scattered participants made possible by quantum-enhanced federated training offer improved user experiences and customization of adaptive content  \cite{park2024joint}. Crucially for trust-building and user privacy in virtual environments, using quantum cryptography techniques such as QKD and BQC guarantees safe, tamper-proof user interactions. But the latency-sensitive character of the Metaverse puts strict criteria on quantum communication infrastructure, which calls for creativity to lower transmission delays and decoherence. Notwithstanding present quantum technological constraints, developments in quantum internet prototypes and quantum error-reducing techniques point to a feasible route towards useful QFL incorporation into Metaverse platforms. Adoption of QFL can ultimately transform Metaverse applications by offering highly immersive, safe, and interactive worlds at hitherto unheard-of computing sizes \cite{ren2022quantum}.
     \item By offering a safe, dispersed framework for threat identification and mitigation across distributed systems, QFL greatly increases network security. Scalability, data privacy, and vulnerability to targeted cyberattacks pose difficulties for conventional centralized solutions to network security \cite{yamany2021oqfl}. Using quantum-enhanced federated algorithms, QFL lets distributed nodes cooperatively identify anomalies and hostile behavior in real-time without disclosing private security data. Utilizing improved pattern recognition and anomaly detection algorithms functioning outside of conventional computational boundaries, quantum computing inherent capabilities improve the identification of complex cyber threats \cite{subramanian2024hybrid}. Furthermore, providing intrinsically safe routes for transmitting important security-related model updates, quantum-secured communication systems, including QKD, helps to prevent interception and guarantee data integrity. Further ensuring confidentiality, QDP lets nodes safely distribute model parameters without violating private operational data \cite{ren2024variational}. Still difficult, nonetheless, actual integration of QFL into network security is hampered by noisy quantum technology, restricted scalability, and quantum resource availability. QFL's ability to transform network security becomes more feasible as quantum technologies develop, enabling more stable quantum hardware and larger-scale quantum network infrastructures, therefore providing durable, privacy-preserving defense mechanisms against developing threats.
 \end{itemize}

\color{black}
Across all these promising applications, a common set of \textbf{practical considerations} and \textbf{unresolved challenges emerges}, primarily centered on hardware readiness and communication infrastructure. For domains like satellite and vehicular networks, the physical deployment of fragile, environmentally sensitive quantum hardware remains a monumental hurdle. This highlights a crucial \textbf{trade-off}: the pursuit of quantum-enhanced performance versus the immense overhead and cost of developing and maintaining the necessary quantum infrastructure. A significant \textbf{unresolved challenge} is the absence of a mature ``Quantum Internet" capable of reliably distributing entanglement over long distances, which is a prerequisite for many advanced QFL protocols. These \textbf{emerging communication constraints} such as high decoherence rates in optical fiber and free space—mean that the performance of QFL in practice is currently bottlenecked by the physical transmission of quantum states, not just by quantum computation. Consequently, a key \textbf{practical consideration} for any real-world QFL application is the co-design of quantum algorithms with the noisy, resource-constrained hardware and networks available in the NISQ era.
\color{black}

\section{Frameworks, Platforms and Prototype Implementations, and A Case Study on QFL} \label{casestudy}

\subsection{Frameworks and Platforms Related to QFL}
\subsubsection{Quantum Programming Frameworks}
\begin{itemize}
  \item \textit{Qiskit (IBM):} Qiskit is an open-source software development kit (SDK) developed by IBM for building, simulating, and executing quantum circuits \cite{wille2019ibm}. It supports simulation through Qiskit Aer and real hardware access via IBM Quantum cloud. It’s widely used for prototyping quantum machine learning and QFL models due to its modular structure and active community.
  \item \textit{PennyLane (Xanadu):} PennyLane is a Python library designed for hybrid quantum-classical machine learning, enabling integration with classical frameworks like PyTorch and TensorFlow \cite{bergholm2018pennylane}. It supports automatic differentiation of quantum circuits, making it ideal for QFL, where parameter updates are based on gradients. PennyLane is particularly strong for implementing variational quantum algorithms and hybrid training loops.
  \item \textit{Cirq (Google):} Cirq is Google’s quantum SDK tailored for NISQ devices, allowing fine-grained control over circuit construction \cite{wang2024toward}. It includes tools for noise modeling and performance tuning, which are essential for QFL testing. Cirq is often used for developing quantum workloads compatible with Google’s quantum hardware.
  \item \textit{Amazon Braket SDK:} Amazon Braket SDK offers a unified interface for designing and running quantum circuits across different quantum processing unit (QPU) providers (IonQ, Rigetti, Oxford Quantum Circuits (OQC)) \cite{markidis2023programming}. It supports both local simulation and cloud deployment, making it a flexible tool for QFL development. Its integration with Amazon Web Services (AWS) facilitates scalability and orchestration in distributed quantum systems.
  \item \textit{Strawberry Fields (Xanadu):} Strawberry Fields is a Python library for photonic quantum computing based on continuous-variable systems \cite{rueda2021continuous}. It supports the simulation of quantum optical circuits and integrates with PennyLane for hybrid training. This tool is especially valuable for exploring QFL on non-qubit-based architectures like photonics.
\end{itemize}

\subsubsection{FL Libraries}

\begin{itemize}
  \item \textit{PyTorch:} PyTorch is a popular deep learning framework known for its flexibility and dynamic computational graphs \cite{ansel2024pytorch}. It integrates seamlessly with PennyLane, enabling hybrid training where quantum circuits act as layers within classical NNs. PyTorch is widely used in QFL to perform client-side training and model updates.
  \item \textit{TensorFlow:} TensorFlow is a scalable machine learning framework developed by Google, offering both eager and graph-based execution modes \cite{agrawal2019tensorflow}. It supports large-scale training and integrates well with FL tools like TensorFlow Federated (TFF). TensorFlow can also be used for hybrid QFL when combined with quantum frameworks.
  \item \textit{Flower (FLWR):} Flower is a modular and extensible federated learning framework that supports backends like PyTorch and TensorFlow \cite{beutel2022flower}. It allows easy implementation of server-client communication, aggregation logic, and global round orchestration in FL or QFL setups. Flower is great for quickly prototyping FL systems across distributed nodes.
  \item \textit{TensorFlow Federated (TFF):} TFF is built on TensorFlow and allows the simulation and deployment of FL algorithms \cite{luzon2024tutorial}. It provides abstractions for client and server logic, helping manage data locality and secure parameter sharing. TFF can be extended for use in QFL by adapting quantum model training into its federated workflow.
  \item \textit{FedML:} FedML is an open-source framework built for large-scale FL across edge, mobile, and cloud environments \cite{chen2023openfed}. It supports cross-device communication, heterogeneous data environments, and flexible training pipelines. FedML's scalability and modularity make it a strong candidate for distributed QFL systems.
\end{itemize}

\subsubsection{Quantum Hardware Backends}

\begin{itemize}
  \item \textit{IBM Quantum:} IBM Quantum provides cloud-based access to superconducting quantum processors through the IBM Q Experience \cite{abughanem2024ibm}. It is integrated with Qiskit, allowing researchers to easily deploy quantum circuits on real hardware. This backend is essential for evaluating QFL models beyond simulation.
  \item \textit{Amazon Braket:} Amazon Braket connects developers to multiple quantum hardware providers, such as IonQ and Rigetti, through a unified interface \cite{bansal2024qualitative}. It supports both real-device execution and simulation, enabling hybrid and scalable QFL deployments. Braket is particularly suited for testing QFL across various hardware types.
  \item \textit{Azure Quantum:} Azure Quantum is Microsoft’s quantum computing platform offering access to multiple QPUs and simulators via the Azure cloud \cite{prateek2023quantum}. It supports hybrid quantum workflows and integrates with classical ML and data services. Azure Quantum is a good fit for QFL experiments in enterprise or cloud-native settings.
  \item \textit{Xanadu Cloud:} Xanadu Cloud offers access to photonic quantum processors and supports continuous-variable quantum computing \cite{abughanem2024nisq}. It integrates seamlessly with the PennyLane ecosystem, enabling hybrid and photonic QFL applications. Xanadu Cloud is one of the few platforms providing non-qubit-based hardware access.
  \item \textit{D-Wave:} D-Wave is a quantum programming framework intended for addressing optimization problems through quantum annealing \cite{bozejko2024optimal}. It offers users tools such as Ocean SDK to articulate issues as quadratic unconstrained binary optimization (QUBO) models and implement them on D-Wave’s quantum hardware that integrates classical and quantum resources, rendering it appropriate for practical applications.
\end{itemize}

\subsubsection{Quantum Simulators}

\begin{itemize}
  \item \textit{Qiskit Aer:} Qiskit Aer is IBM’s high-performance simulator for executing quantum circuits with various noise models \cite{faj2023quantum}. It enables developers to debug QFL logic and analyze performance before running on real hardware. Aer supports multiple simulation backends like statevector and QASM simulators.
  \item \textit{PennyLane Simulators:} PennyLane includes built-in simulators such as \texttt{default.qubit} that support hybrid quantum-classical models \cite{zappin2024quantum}. These simulators allow for gradient computation and circuit differentiation, which is key in training QFL models. They are perfect for local experimentation before moving to hardware.
  \item \textit{Cirq Simulator:} The Cirq simulator mimics the behavior of NISQ devices, enabling testing of quantum circuits with realistic noise and gate configurations \cite{li2023qasmbench}. It helps researchers prepare QFL models for deployment on Google’s quantum processors. The simulator supports fine-grained diagnostics and parameter tracking.
  \item \textit{Braket Local Simulator:} The Braket Local Simulator mimics the behavior of hardware supported on the Amazon Braket platform \cite{valencia2021hybrid}. It is useful for verifying QFL circuits and workflows before committing to expensive real-device execution. This tool supports rapid iteration and debugging in a local environment.
\end{itemize}

\subsection{Prototype Implementations on QFL}
\begin{itemize}
    \item \textit{QFL for Liver Biopsy Analysis:} This study combines QML and FL to classify liver biopsy images, specifically diagnosing hepatic steatosis \cite{lusnig2024hybrid}. By creating a dataset of 4,400 samples across four steatosis stages, the research aims to surpass the diagnostic accuracy of human experts. The FL approach enables collaborative training across multiple clients without compromising data privacy, adhering to GDPR standards. Cite this: Hybrid Quantum Image Classification and FL for Hepatic Steatosis Diagnosis
    \item \textit{QFL for Medical Image Classification:} This research introduces a federated hybrid quantum–classical algorithm called the quanvolutional neural network (QNN), designed for collaborative learning across healthcare institutions without sharing data \cite{bhatia2023federated}. The study evaluates the model's performance on medical datasets, including COVID-19 and MedNIST, demonstrating its robustness and feasibility in handling non-IID and real-world data distributions. The findings indicate a significant reduction in communication rounds compared to traditional federated stochastic gradient descent approaches.
     \item \textit{QFL for Financial Fraud Detection:} This research presents QFNN-FFD, a framework that merges QML with FL to enhance financial fraud detection \cite{innan2024qfnn}. By leveraging quantum computing's computational power and FL's data privacy features, QFNN-FFD offers a secure and efficient method for identifying fraudulent transactions.
\end{itemize}

\subsection{A Case Study}
We present an example from our research to show how QFL can be useful in the real world. In this case, we used QFL to find unusual patterns (anomalies) in a system to see if using a quantum-based approach can enhance the detection accuracy of anomalies. 

\subsubsection{Overview}
\begin{figure}
    \centering
    \includegraphics[width=0.99\linewidth]{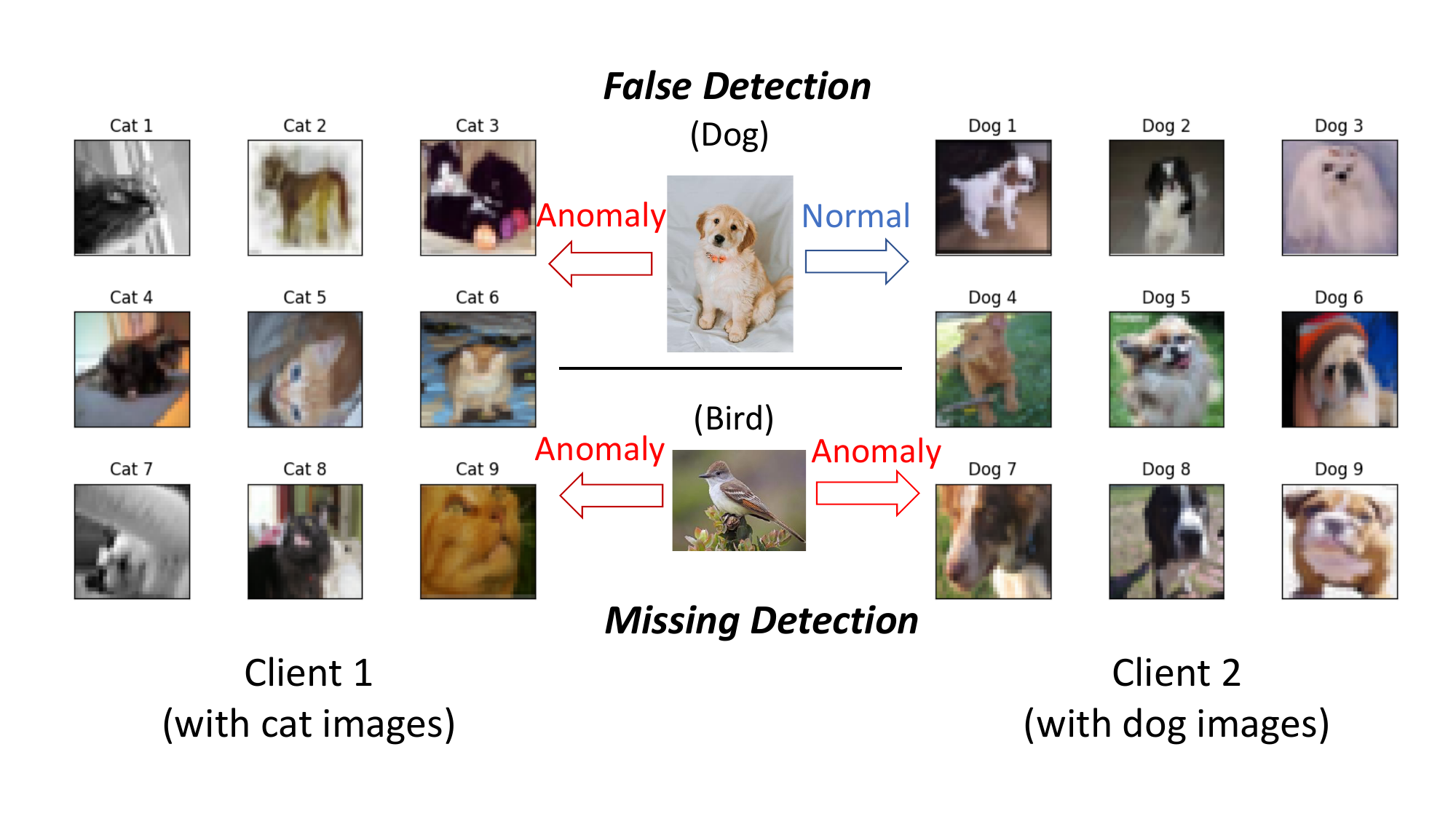}
    \caption{An example of false and missed detection issues in QFL anomaly detection on CIFAR-10 data. In this case, cat and dog photographs are initially regarded as normal. False detection occurs when a dog image is considered a cat, and missing detection occurs when a bird is considered a cat or a dog. }
    \label{fig: anomaly}
\end{figure}

Detecting unusual patterns is essential for keeping today’s smart systems secure and running smoothly. These irregularities can point to things like fraud, equipment faults, or cybersecurity attacks illustrate in Fig. ~\ref{fig: anomaly}. For the traditional approach, there are three main types of anomaly detection techniques—reconstruction-based (RE) \cite{zhou2017anomaly, wen2021feddetect,rahman2024multimodal}, classification-based (CF) \cite{lane1997application,nguyen2019scalable, ruff2018deep, reiss2021panda}, and constructive learning (CL) approaches \cite{tack2020csi,han2021elsa,dong2024fadngs}. 
Although traditional machine learning methods are often used for detecting anomalies, the detection of anomalies has some unique problems and challenges. i) Acquiring vast amounts of abnormal data is naturally challenging. ii) The distinctions between regular and abnormal events are sometimes extremely small in size. iii) Specific learning approaches and cost-sensitive solutions are frequently necessary to address the data imbalance, increasing computational and design complexity. iv) Finally, the definition of an anomaly varies greatly depending on the context. A pattern that appears abnormal in one area or system may be perfectly acceptable in another. 

\begin{figure*}
    \centering
    \footnotesize
    \begin{subfigure}[t]{0.45\linewidth} 
        \centering
        \includegraphics[width=\linewidth]{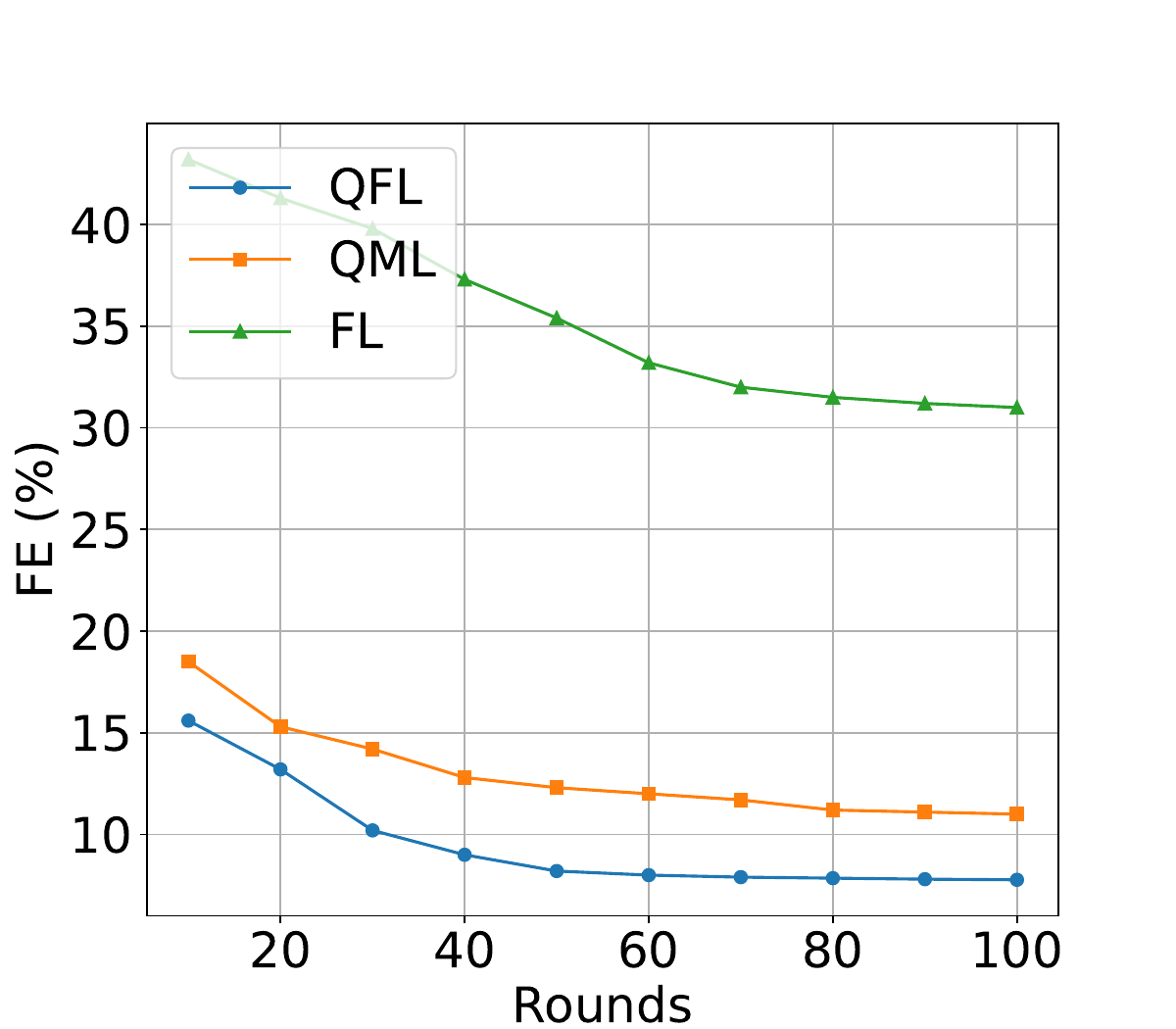}
        \caption{FE (\%)}
        \label{fig1a}
    \end{subfigure}
    \hfill 
    \begin{subfigure}[t]{0.45\linewidth} 
        \centering
        \includegraphics[width=\linewidth]{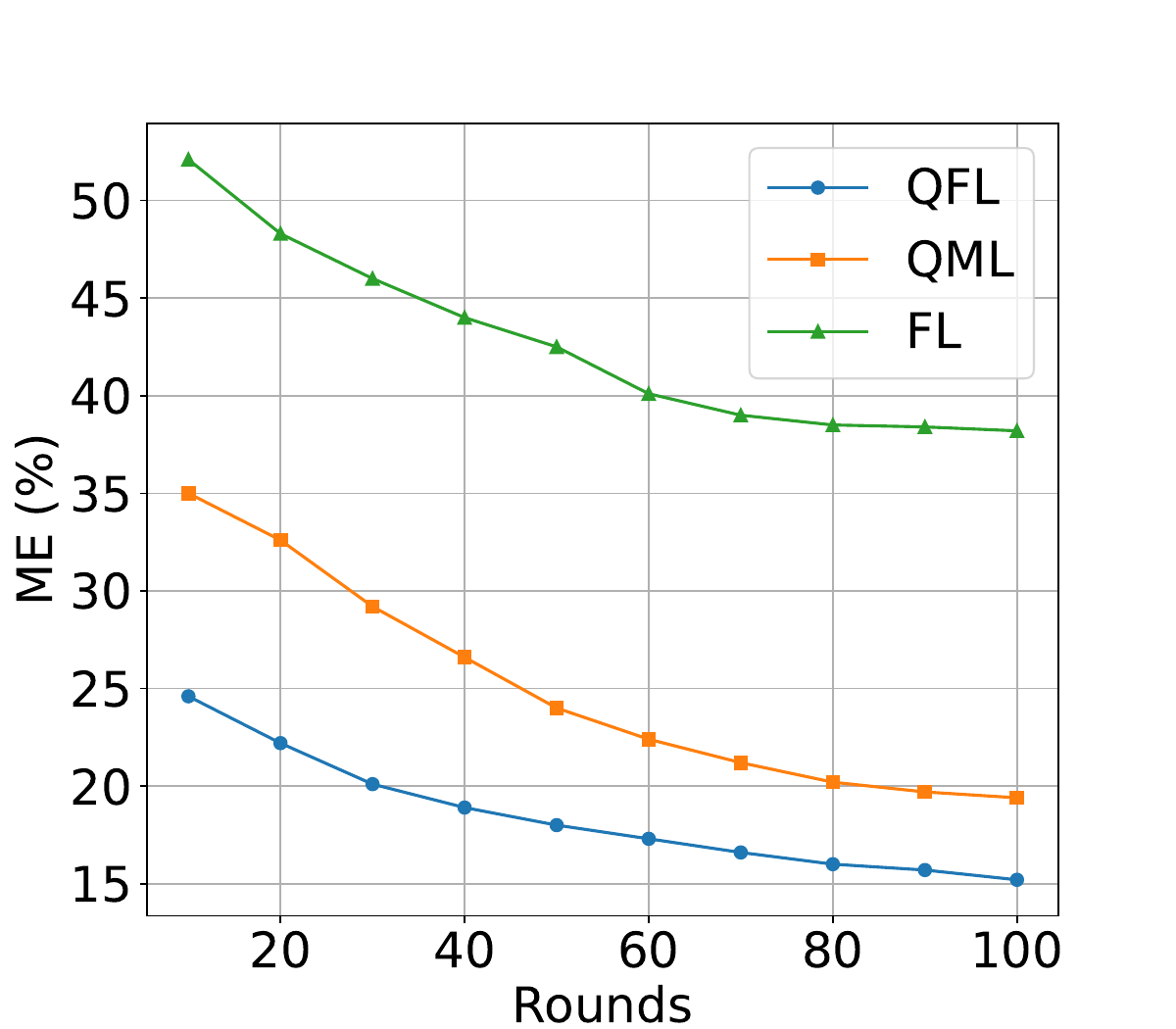}
        \caption{ME (\%)}
        \label{fig1b}
    \end{subfigure}
    \vspace{-5pt} 
    \caption{Training performance comparison between QFL, QNN, and FL approaches. we compared both FE (\%) (a) and ME (\%) (b) in anomaly detection where we use CIFAR-10 as normal dataset and CIFAR-100 as anomaly dataset. }
    \label{fig1}
\end{figure*}

QML can surpass traditional ML techniques in anomaly detection by overcoming fundamental constraints with the unique features of quantum. QML allows data to be encoded into high-dimensional quantum spaces, allowing models to learn efficiently even from sparse anomalous samples. 
For example, \cite{mangini2022quantum,kukliansky2024network} presented a QNN method that enables feature extraction and dimensionality reduction at the same time for industrial quality control by combining a classifier and an autoencoder. \cite{wang2022data} used QNNs to recover damaged or missing data, emphasizing the significance of precise quantum state preparation. Despite these encouraging research efforts, QML presents some limitations. The traditional method in QML of using a single server to handle data from several sources raises data privacy issues. Data centralization in QML is particularly dangerous for applications that require private or sensitive data since it leaves the data vulnerable to breaches either in transit or on the central server, which might lead to data-related attacks on the system. Furthermore, QML requires substantial quantum processing power to carry out computations on the central device. Quantum resources, including coherent quantum gates and quantum bits, are still in the early stages of research and can be costly and challenging to sustain over time.

As a result, we provide an anomaly detection approach based on Quantum Federated Learning (QFL) to overcome these difficulties.  QFL combines the power of quantum computing with the privacy benefits of federated learning, allowing numerous clients to train models simultaneously without sharing raw data.  This strategy eliminates the need for central data gathering, lowers privacy risks, and promotes learning from decentralized, non-IID data.  Simultaneously, quantum improvements improve the accuracy with which unusual and subtle abnormalities are represented.  By spreading quantum learning across clients, QFL eliminates the need for centralized quantum resources, addressing privacy and scalability problems while enhancing anomaly detection performance.

\subsubsection{Proposed Method} 
Our approach to anomaly detection addresses two common problems: false and missed detections, as discussed in \cite{dong2024fadngs}. A false detection occurs when something normal, such as a regular image of a dog, is incorrectly identified as an anomaly (it is not a dog). This frequently occurs because each client (device or system) has different types of data, which can be confusing. On the other hand, a missed detection occurs when a genuine anomaly, such as an image of a bird, goes unnoticed and is identified as a dog, which generally occurs because the client lacked sufficient similar data to recognize it as an anomaly. To address these issues, we first set up a shared reference (a global density function) that allows all clients to agree on what is "normal" and what is not, and we add some noise to keep data private. Second, we use contrastive learning, a type of training that teaches the system to bring similar (normal) items closer together while pushing different (anomalous) items apart. Finally, rather than simply combining model weights, we employ an ensemble distillation strategy that focuses on how well local models detect anomalies, applying their strengths to guide the training of a more accurate global model.

We investigate a hybrid quantum-classical approach, more specifically VQE, for anomaly detection. The proposed QFL framework is designed around a central server and $n$ dispersed clients. Each client has its private data for finding unusual patterns. They train a model using just their data and never share the raw information, so privacy is protected. The central server doesn’t have access to the data—it just collects the results from each client’s training, combines them to make a better overall model, and sends that back to the clients. This cycle keeps repeating, allowing everyone to learn together without sharing private data.

Each client trains a local anomaly detection model with a QNN based on PQCs. This procedure starts with encoding classical input characteristics into quantum states using methods like angle or amplitude encoding, which allows the data to be represented in a quantum Hilbert space. Once encoded, the quantum states are sent via a series of parameterized quantum gates that comprise the PQC. These gates determine the model's learnable structure, and their parameters change throughout training. After the circuit, quantum measurements turn the quantum state back into classical information, which is then utilized to compute the loss and change the parameters using gradient-based algorithms. 
After local training, clients simply send their PQC parameters to the central server; no data or quantum states are transferred. The server aggregates model data via a federated averaging approach, which integrates parameter sets from all clients into a single global model. This revised model gathers information from various data distributions and broadcasts it to clients to start a new training round. This cycle repeats for numerous global rounds before the model stabilizes.

\begin{table*}[htbp]
\footnotesize
\centering
\begin{tabular}{@{}|p{1.4cm}|p{1.6cm}|c|c|c|c|c|c|@{}}
\toprule
Normal Dataset & Anomaly Dataset & \multicolumn{2}{c|}{Method} & FE (\%) $\downarrow$ & ME (\%) $\downarrow$ &AUROC (\%) $\uparrow$ & AUPR (\%) $\uparrow$ \\
\midrule
\multirow{18}{*}{CIFAR-10} & \multirow{9}{*}{CIFAR-100} & \multirow{2}{*}{RE} & AE \cite{zhou2017anomaly} & 45.6 $\pm$ 0.8 & 48.9 $\pm$ 1.2 & 55.8 $\pm$ 0.9 & 56.0 $\pm$ 2.0 \\
                                 &&& FedDetect \cite{wen2021feddetect} & 46.2 $\pm$ 1.2 & 54.9 $\pm$ 0.9 & 56.4 $\pm$ 0.9                        & 55.9 $\pm$ 0.9                      \\
                                 \cline{3-8}
                                &  & \multirow{3}{*}{CF}      & OCSVM \cite{nguyen2019scalable}     & 46.2 $\pm$ 1.8                       & 47.3 $\pm$ 1.9                       & 54.8 $\pm$ 1.2                        & 55.2 $\pm$ 1.3                      \\
                                 &      &    & Deep-SVDD \cite{ruff2018deep} & 48.6 $\pm$ 1.2                       & 46.4 $\pm$ 0.8                       & 53.9 $\pm$ 1.5                        & 56.2 $\pm$ 1.2                      \\
                                 &      &    & Panda \cite{reiss2021panda}     & 47.6 $\pm$ 0.7                       & 46.3 $\pm$ 1.2                       & 55.5 $\pm$ 1.5                        & 54.2 $\pm$ 0.4                      \\
                                 \cline{3-8}
                                 & & \multirow{4}{*}{CL}       & CSI \cite{tack2020csi}       & 43.2 $\pm$ 0.8                       & 42.8 $\pm$ 2.0                       & 58.6 $\pm$ 1.1                        & 61.1 $\pm$ 0.8                      \\
                                 &      &    & Elsa \cite{han2021elsa}      & 46.2 $\pm$ 1.2                       & 40.1 $\pm$ 2.7                       & 54.0 $\pm$ 1.0                        & 56.3 $\pm$ 2.7                      \\
                                 &     &     & FADngs \cite{dong2024fadngs}         & 30.6 $\pm$ 1.0                       & 25.0 $\pm$ 0.6                       & 77.8 $\pm$ 0.9                        & 75.1 $\pm$ 0.9                      \\ 
                                 &     &     & QNN \cite{kukliansky2024network}         & 20.5 $\pm$ 1.4                       & 19.0 $\pm$ 1.7                       & 81.0 $\pm$ 0.8                        & 85.6 $\pm$ 1.1                      \\ 
                                &     &     & QFL & \textbf{7.77 $\pm$ 1.8}                       & \textbf{15.2 $\pm$ 1.0}                       & \textbf{85.2 $\pm$ 2.1}                        & \textbf{94.1 $\pm$ 2.2}      \\\cline{2-8}
 & \multirow{9}{*}{\shortstack{Tiny-\\ImageNet}}       & \multirow{2}{*}{RE}   & AE \cite{zhou2017anomaly} & 48.0 $\pm$ 0.8 & 40.9 $\pm$ 0.8 & 55.1 $\pm$ 1.4 & 52.6 $\pm$ 0.9 \\
                                 &     &     & FedDetect \cite{wen2021feddetect} & 48.1 $\pm$ 1.8 & 53.9 $\pm$ 1.0 & 55.9 $\pm$ 1.0 & 53.03 $\pm$ 1.0 \\ 
                                 \cline{3-8}
                                &  & \multirow{3}{*}{CF}      & OCSVM \cite{nguyen2019scalable} & 48.4 $\pm$ 1.6 & 47.7 $\pm$ 3.0 & 52.5 $\pm$ 1.5 & 51.8 $\pm$ 1.6 \\
                                 &      &    & Deep-SVDD \cite{ruff2018deep} & 48.2 $\pm$ 0.8 & 48.2 $\pm$ 1.4 & 52.6 $\pm$ 2.2 & 51.4 $\pm$ 0.9 \\
                                 &      &    & Panda \cite{reiss2021panda} & 47.8 $\pm$ 0.4 & 48.5 $\pm$ 1.5 & 52.2 $\pm$ 2.0 & 51.6 $\pm$ 0.6 \\
                                 \cline{3-8}
                                 & & \multirow{4}{*}{CL}       &CSI \cite{tack2020csi} & 43.6 $\pm$ 0.6 & 42.0 $\pm$ 3.3 & 59.1 $\pm$ 0.9 & 60.4 $\pm$ 0.8 \\
                                 &      &    & Elsa \cite{han2021elsa} & 47.2 $\pm$ 1.4 & 47.7 $\pm$ 1.7 & 54.4 $\pm$ 0.5 & 53.8 $\pm$ 1.7 \\
                                  &     &     & FADngs \cite{dong2024fadngs}         & 25.1 $\pm$ 1.0& 20.2 $\pm$ 2.3& 83.8 $\pm$ 0.9& 82.8 $\pm$ 1.1\\ 
                                  &     &     & QNN \cite{kukliansky2024network}         & 15.9 $\pm$ 0.9                       & 18.7 $\pm$ 1.6                       & 81.8 $\pm$ 0.9                        & 87.3 $\pm$ 2.9                      \\ 
                                &     &     & QFL & \textbf{10.6 $\pm$ 1.7}                       & \textbf{12.5 $\pm$ 2.9}                       & \textbf{85.9 $\pm$ 2.1}                        & \textbf{91.2 $\pm$ 1.7}                     \\ \bottomrule
\end{tabular}
\caption{A Comparative analysis of anomaly detection methods across CIFAR-100 and Tiny-ImageNet datasets, showing performance metrics such as False Error (FE), Missing Error (ME), AUROC, and AUPR for various methods.}
\label{Table: stateoftheart}
\end{table*}
\subsubsection{Illustrative Results}
To evaluate the performance of our QFL-based approach, to represent normal data, we use the baseline dataset CIFAR-10 \cite{krizhevsky2009learning}. This dataset contains a diverse but uniformly labeled collection of images across ten classes. For anomaly data, we use CIFAR-100 \cite{krizhevsky2009learning} and Tiny-ImageNet \cite{deng2009imagenet}. CIFAR-100, with its 100 classes, offers greater diversity, while Tiny-Imagenet, with its larger dimensions, provides a more detailed comparison. We employ two common anomaly detection metrics, AUROC (Area Under Receiver Operating Characteristic Curve) \cite{bradley1997use} and AUPR (Area Under Precision-Recall Curve) \cite{boyd2013area}, where higher values indicate better performance, as well as False Error (FE), which calculates the rate of normal samples that are mistakenly classified as anomalies, and Missing Error (ME), which calculates the rate of actual anomalies missed.

First, to show the effectiveness of QFL, we compare QFL with QNN and FL in Fig.~\ref{fig1} and compare using FE and ME. The figure shows that
QFL consistently performs better than both QNN and FL, staying on top across all training rounds—even though its scores slowly go down over time. For example, QNN performance drops from 18.5 to 11.0 in one case (FE) and from 35.0 to 19.4 in another (ME). QFL also drops, but it starts lower and still ends up with the best results—going from 15.6 to 7.77 in FE and from 24.6 to 15.2 in ME. In short, QFL gives the most accurate results overall.

Finally, A thorough performance comparison between multiple anomaly detection techniques evaluated in two settings—CIFAR-10 (normal data) against CIFAR-100 and Tiny-ImageNet (anomalous data)—is shown in Table~\ref{Table: stateoftheart}. Traditional RE techniques, such as FedDetect \cite{wen2021feddetect} and AE \cite{zhou2017anomaly}, display greater FE and ME rates in both cases, suggesting that they are not very robust in identifying abnormalities in decentralized systems. Similarly, CF techniques like OCSVM \cite{nguyen2019scalable}, Deep-SVDD \cite{ruff2018deep}, and Panda \cite{reiss2021panda} exhibit slight gains but continue to have issues with recall and detection precision. Recent CL techniques such as CSI \cite{tack2020csi} and Elsa \cite{han2021elsa} demonstrate a better balance between false and missing errors, while FADngs \cite{dong2024fadngs} federated contrastive learning design helps it achieve significant improvements in AUROC and AUPR. Our suggested QFL approach, however, performs best on all measures, showing the best AUROC and AUPR values in all settings and noticeably reduced false and missing error rates. When comparing CIFAR-10 to CIFAR-100, QFL outperforms even the quantum baseline QNN, lowering the False Error to 7.77\%, the Missing Error to 15.2\%, and achieving an AUROC of 85.2\% and AUPR of 94.1\%. QFL continues to perform well in the more complicated CIFAR-10 vs. Tiny-ImageNet scenario, with the greatest AUROC (85.9\%) and AUPR (91.2\%) along with FE of 10.6\% and ME of 12.5\%. These findings show that QFL is a reliable and privacy-preserving solution for federated systems as it not only retains good anomaly detection performance but also successfully adjusts to various anomaly distributions. This also highlights QFL's ability to handle complex, multidimensional, and context-dependent data while maintaining data privacy. As quantum hardware advances, QFL has the potential to be a fundamental method in safe and intelligent decentralized solutions to real-life problems.

\textcolor{black}{This case study provides a concrete example of how the theoretical elements of QFL can be integrated using current platforms to deliver tangible performance and security benefits. While such demonstrations are crucial for validating the promise of QFL, they also highlight open issues. Building upon these practical insights, the final section of our survey will now broaden the scope to address the key unresolved challenges, open research questions, and exciting future directions that will shape the next phase of Quantum Federated Learning.}

\section {Challenges and Future Directions} \label{challengesandfuturedirections}

In this section, we discuss a few important research challenges and potential solutions toward efficient QFL. 
\subsection{System Heterogeneity in QFL}
\textcolor{black}{System heterogeneity is a major challenge due to the variability in the quantum capabilities of edge devices. One key source of this heterogeneity is the difference in the number of qubits available on each participating device. Each quantum-powered edge device may have access to a different number of qubits, which influences the complexity of QML models that can run. Devices with fewer qubits are constrained in the size and complexity of the quantum circuits they can process, while those with more qubits can handle larger, more intricate models. This difference in quantum resources complicates the process of sharing model updates across devices, as the models trained on each device may differ significantly in complexity and size \cite{nawaz2019quantum, granelli2022novel}.}

Another aspect of heterogeneity in QFL arises from the variation in parameterized gate layers between devices. Quantum circuits used for QML typically consist of multiple layers of parameterized gates, but not all quantum devices are capable of supporting the same level of complexity in these circuits. Some devices can implement deeper or more complex gate layers, offering greater flexibility in how quantum states are represented and optimized, while others may be limited in their gate depth. These disparities in gate layer capabilities create inconsistencies in the models trained on different devices, making it difficult to aggregate the model updates in a standardized way. To address this, advanced techniques are required to ensure that the model updates can be effectively integrated and the training process remains consistent across a heterogeneous network of quantum devices \cite{tsang2023hybrid}. In terms of designing adaptive QFL algorithms that dynamically change learning rates, communication intervals, or circuit complexity to fit each node hardware limits and fidelity levels is part of future directions. 


\begin{table*}
\centering
\caption{Summary of Research Challenges, Solution, and Direction for QFL.}
\label{table:challenges}
\rowcolors{2}{gray!10}{white} 
\begin{tabular}{m{3.3cm}|m{4.7cm}|m{4cm}|m{4cm}}
\hline
\rowcolor{gray!30}
\makecell[c]{\textbf{Chllenges}} & \makecell[c]{\textbf{Description}} & \makecell[c]{\textbf{Solutions}} & \makecell[c]{\textbf{Possible direction}} \\ \hline
System Heterogeneity in QFL & 
\begin{itemize}
    \item Devices differ in qubit count, impacting the complexity and size of feasible QML models.
\end{itemize}
& 
\begin{itemize}
    \item Implement advanced integration techniques to harmonize model updates across devices with varying quantum circuit complexities\cite{tsang2023hybrid}.
\end{itemize}
&
\begin{itemize}
    \item Explore adaptive algorithms and federated transfer learning to enhance model compatibility and learning efficiency \cite{bilal2024bc,nawaz2019quantum}.
\end{itemize}
 \\ \hline
Quantum Noise Control in NISQ-based QFL & 
\begin{itemize}
    \item NISQ devices, characterized by limited qubit counts and short coherence periods, struggle with decoherence, gate errors, and measurement inaccuracies.
\end{itemize} &
\begin{itemize}
    \item Employ error mitigation techniques like zero-noise extrapolation and probabilistic error canceling, though they require multiple circuit runs \cite{filippov2024scalability}.
\end{itemize}
&\begin{itemize}
   \item Develop noise-aware QFL algorithms that adjust learning parameters based on real-time device performance, and explore hybrid error-correcting approaches to enhance scalability and efficiency in QFL systems \cite{arya2023quantum, nawaz2019quantum}.
\end{itemize} \\ \hline
Standardization in QFL  & 
\begin{itemize}
    \item As QFL moves towards real-world application, establishing industry-wide standards for interoperability, security, and reliability is crucial.
\end{itemize} &
\begin{itemize}
    \item Develop global standards including performance criteria for quantum circuits, data formats, and secure communication protocols \cite{kazmi2024security}.
\end{itemize}
&\begin{itemize}
    \item Foster international regulatory cooperation and create supportive policies for QFL innovation, ensuring safe and efficient system deployment across critical sectors \cite{abughanem2024nisq, kannan2024quantum, roy2024standardization}.
\end{itemize} \\ \hline
Integration with 6G Networks & 
\begin{itemize}
    \item QFL must align with 6G's ultra-low latency, massive connectivity, and scalability through effective quantum-safe communications and resource management.
\end{itemize} &
\begin{itemize}
    \item  Implement advanced quantum repeaters, error-correcting codes, and network slicing for efficient quantum state transfer and resource allocation \cite{blika2024federated, dutta2024quantum}.
\end{itemize}
&\begin{itemize}
    \item Develop hybrid networking solutions that integrate quantum and classical layers, ensuring seamless operation within 6G frameworks \cite{kannan2024revolutionizing, alhaj2023integration}.
\end{itemize} \\ \hline
\end{tabular}
\end{table*}

\subsection {Quantum Noise Control in NISQ-based QFL}

Control of quantum noise is a major obstacle for QFL in Noisy Intermediate-Scale Quantum (NISQ) systems. These devices are prone to decoherence, gate mistakes, and measurement errors, as by definition their qubit counts and coherence periods are limited \cite{arya2023quantum, khalid2024quantum}. In QFL settings—where several quantum nodes collaboratively train a global model utilizing PQCs—noise can accumulate unpredictably across various nodes, creating uneven error rates and perhaps skewing federated model updates. Because gate fidelities and noise levels differ between customers, this heterogeneity sometimes results in slower convergence or less-than-ideal performance.

Additionally, an important limitation is the low viability of full-scale QEC in current NISQ systems. Rather, to somewhat offset the consequences of noise, researchers depend on error-mitigating techniques, including zero-noise extrapolation and probabilistic error canceling. But such methods can call for repeated circuit runs—an overhead that would not be feasible for geographically scattered or resource-limited quantum devices in a QFL system. Furthermore, it is still difficult to create QFL protocols that balance local training stages with communication on noisy channels since conventional federated averaging might not consider important error variations between nodes \cite{filippov2024scalability}.

Future initiatives include the development of noise-aware QFL algorithms that dynamically change the learning rate or depth of quantum circuits depending on real-time device authenticity measurements. Using hybrid error-correcting techniques—that is, partial QEC or quantum error detection codes—to stop critical error propagation without the complete expense of fault-tolerance offers still another interesting path. Furthermore, dynamically arranging activities across nodes with different noise levels—possibly via flexible scheduling or quantum resource pooling—may improve dependability and performance in large-scale installations \cite{nawaz2019quantum}. Together with continuous hardware advances, these techniques represent important milestones toward a strong and scalable NISQ-based QFL.




\subsection{Standardization}
Establishing industry-wide standards and regulatory frameworks becomes ever more important as QFL advances from conceptual development toward practical implementation. Consistent rules are needed to guarantee interoperability, security, and dependability across heterogeneous platforms since QFL uses distributed learning topologies, quantum computing capabilities, and advanced cryptographic protocols like QKD \cite{kazmi2024security}. Such guidelines would include issues including universal performance criteria for PQCs, data interchange formats between quantum and classical systems, and protocol definitions for quantum-safe communications.

Apart from technical harmonization, government control and policy support will be essential to create an environment fit for the development of QFL software. While preserving robust privacy protections and security guarantees, policies supporting public-private partnerships and investments in quantum technology R\&D could speed innovation \cite{abughanem2024nisq}. Regulatory authorities may also specify compliance strategies for quantum data privacy, thereby ensuring that QFL systems follow strict guidelines before being used in national defense, banking, and crucial industries, including healthcare. Transparency in data governance and model update policies will assist authorities in reducing hazards and safeguarding user-sensitive data in major quantum installations \cite{kannan2024quantum}.

Furthermore, ensuring cross-border compatibility and accelerating information flow is the establishment of internationally accepted standards, probably through consortia such as the Institute of Electrical and Electronics Engineers and the International Organization for Standardization. This worldwide approach to control and standard-setting will be vital since QFL development sometimes entails cooperation across several areas, each with its own legal systems. Ultimately, the safe, secure, and scalable adoption of QFL depends on uniform standards and prudent government involvement so enabling different stakeholders to trust, invest in, and gain from quantum-enhanced FL systems \cite{roy2024standardization}.


\subsection {Integration with 6G Networks}
Given the stringent requirements of 6G networks—such as ultra-low latency, massive connectivity, and high scalability—the integration of QFL into next-generation wireless infrastructures offers significant technological promise alongside substantial challenges \cite{duong2022quantum}. At its core, QFL relies on distributed quantum or hybrid quantum-classical devices that collaboratively train models while preserving data privacy and locality. However, supporting such quantum workloads over 6G environments demands robust architectural designs and communication protocols capable of jointly addressing the constraints of both quantum and classical resources \cite{blika2024federated}. Effective quantum-safe communication is required to meet ultra-low latency and security requirements from end users. However, current quantum channels sometimes suffer from restricted range and decoherence. Reliable quantum state transfer with minimum delay is challenging since it depends on quantum repeaters, error-correcting codes, or entanglement-assisted routing to preserve coherence across large distances \cite{dutta2024quantum}. Moreover, with 6G predicted to enable vast machine-type communication, scalability challenges are raised by large-scale QFL. The quantum hardware capabilities of every node differ; hence, maintaining safe consensus across several quantum devices generates synchronization and scheduling complexity that requires tailored quantum-aware aggregation techniques \cite{kannan2024revolutionizing}.

Given the high bandwidth consumption of quantum-based protocols and the computational expense of error mitigation on NISQ devices, resource management also becomes crucial. Dynamic allocation of quantum resources in line with sophisticated 6G characteristics, such as network slicing, could help balance load and enable heterogeneous device capabilities. Finally, close interdependent quantum and classical layer interactions hinder protocol design. Without sacrificing security or performance, researchers have to create hybrid networking solutions that effortlessly combine quantum channels with advanced 6G capabilities, including massive multiple-input and multiple-output (MIMO) or terahertz communication \cite{alhaj2023integration}. From creating new quantum network topologies fit for 6G to using standardized quantum-safe techniques, addressing these problems calls for multidisciplinary efforts. Development in these fields will define whether low-latency, large-scale QFL installations in next-generation wireless environments are feasible. We summarize the discussed challenges and solutions in Table~\ref{table:challenges}.


\section {Conclusion} \label{conclusions}
This paper has delved into the opportunities presented by QFL to empower distributed networks through a state-of-the-art survey and in-depth discussions based on recent research in the field. This work is motivated by the absence of a comprehensive survey focusing on all aspects of QFL, from foundational concepts to advanced applications. To bridge this gap, we have first introduced the recent advancements in QFL, followed by its market opportunities and background. We have then analyzed the integration of quantum computing and FL, while investigating its working principle. Subsequently, we have reviewed the fundamentals of QFL and its taxonomy. Specifically, we have explored federation architecture, networking topology, communication schemes, optimization techniques, and security mechanisms within QFL frameworks. Its applications have been also studied, e.g.,  vehicular networks, healthcare networks, satellite networks, metaverse, and network security. From the extensive survey, key insights and lessons learned have also been summarized and analyzed. Finally, we have identified several key research challenges and potential directions for future exploration. We believe that this article will spark greater interest in this emerging field of QFL and inspire future research efforts towards leveraging its full potential.

\balance
\bibliographystyle{ieeetr}
\bibliography{reference}

\end{document}